# Mapping License Plate Recoverability Under Extreme Viewing Angles for Opportunistic Urban Sensing

**Igor Adamenko[1], Orpaz Ben Aharon[1], Yehudit Aperstein[2,*], and Alexander Apartsin[3]**

[1] School of Electrical Engineering, Afeka Academic College of Engineering, Tel Aviv, 69988, Israel; igor.admenko@s.afeka.ac.il, Orpaz.ben.aharon@s.afeka.ac.il

[2] Intelligent Systems, Afeka Academic College of Engineering, Tel Aviv, 69988, Israel; apersteiny@afeka.ac.il

[3] School of Computer Science, Faculty of Sciences, Holon Institute of Technology, Holon, 58102, Israel; alexanderap@hit.ac.il

* Correspondence: apersteiny@afeka.ac.il

**Abstract**

Urban environments are saturated with imaging sensors deployed for purposes unrelated to vehicle identification, from ATM and dashboard cameras to pole-mounted CCTV and smartphones. We term the use of such non-purpose-built sensors for secondary inference "opportunistic sensing"; its central question is where, under uncontrolled capture conditions, AI-enabled restoration remains reliable. This paper introduces *recoverability maps*, a task-agnostic methodology for quantifying that boundary, and applies it to oblique-view license plate recognition (LPR). It pairs a full-grid synthetic sweep of the degradation space with two summary measures: a boundary area-under-curve for coverage and a reliability score $F$ for the frequency and depth of interior unrecovered pockets. For LPR, the space is the oblique-angle grid $[0^\circ, 89^\circ]^2$ sampled by Scrambled Sobol sequences, and the utility is plate-level optical character recognition (OCR) accuracy. Within this synthetic benchmark, approximately $90–92\%$ of the angle grid is recoverable (best single model to union of restoration arms); recovery degrades sharply beyond roughly $80^\circ$ in both axes, and lateral rotations are harder to reconstruct than elevational ones. Five restoration architectures cluster within a narrow AUC band of $0.89–0.93$ and share the same $\alpha/\beta$ asymmetry, so the recoverable region is set primarily by sensing geometry, with architecture affecting efficiency and interior consistency; discriminative architectures outperform generative models. The methodology is validated on real plates: on CCPD and the Brazilian legacy and Mercosur layouts of RodoSol-ALPR, restoration raises held-out extreme-angle recognition by $+15$ to $+38$ exact-match points under plate-specialized recognizers, and the discriminative-over-generative ordering reproduces on real data.

## 1. Introduction

Urban infrastructure routinely accumulates data from sensors installed for one purpose that can be reused for another at near-zero marginal cost. Smartphone accelerometers, originally built for screen-orientation and step counting, have been used to map road-surface defects from ordinary driving [1]. Smart-meter electricity readings, installed for billing, carry enough detail to infer appliance-level activity patterns. Ambient microphones on mobile devices, deployed for voice assistants, can estimate traffic density and crowd activity [2]. Parking-sensor magnetometers and pedestrian-counter beam-breakers similarly admit secondary interpretations beyond their design specification. This pattern, which Campbell et al. [3] and Ganti et al. [4] termed *opportunistic sensing*, has the same underlying structure in every case: a sensor deployed for purpose A produces a signal rich enough to support a secondary inference B, provided B's signal survives the sensor's original acquisition constraints. The practical question is rarely whether opportunistic reuse is possible in principle, but where in the space of uncontrolled acquisition conditions the secondary inference remains reliable.

Imaging sensors are the most information-dense instance of this pattern. Smart cities host over one billion surveillance cameras [5], together with an order of magnitude more mobile cameras on vehicles and handheld devices, almost none calibrated for any specific downstream inference. When a vehicle of interest passes within range of one of these cameras, the captured image can be the only available record. Because the sensor was positioned for an unrelated task, however, the plate typically appears at oblique angles, foreshortened, blurred, and compressed, conditions that place standard license plate recognition (LPR) pipelines outside their usual operating envelope. Modern deep-learning image restoration plausibly recovers such plates, but the angular extent of this recovery has not been mapped. This paper asks: given a camera that cannot be repositioned, at what viewing angles does enough signal survive for automatic LPR to succeed?

Public benchmarks for recognition tasks are typically task-purposed: their captures are curated under controlled or semi-controlled conditions that foreground readability and downstream success. This curation is appropriate when the task is the primary goal of the sensor, but it leaves an inferential gap when the same sensor is used opportunistically for a secondary task. The conditions under which opportunistic sensing operates, namely the extreme corners of the acquisition-parameter space, are systematically underrepresented in benchmarks that were never designed to cover them. A synthetic benchmark that samples the full parameter grid, including the extreme corners, is therefore a natural tool for characterizing the boundary of what opportunistic sensing can recover; its role is not to replace real-world evaluation but to map a regime that real benchmarks do not reach. The resulting numbers should therefore be read as benchmark-bound estimates of an acquisition regime, not as a substitute for field validation on real urban cameras.

Existing LPR research assumes near-frontal capture, typically $|\alpha|, |\beta| \leq 30^\circ$, and degrades rapidly beyond this range [6, 7]. In opportunistic captures, by contrast, the plate routinely appears at $\alpha, \beta > 60^\circ$. The angular extent within which modern deep-learning image restoration can still recover a readable plate has not been mapped.

**Contributions.** The primary contribution of this paper is methodological: we introduce *recoverability maps*, a task-agnostic framework for quantifying where in a parameterized degradation space an opportunistic sensing task remains reliable, together with two summary measures (a boundary area-under-curve and a reliability score $F$) that are adapted from classical operating-characteristic and envelope analyses [8] to the two-dimensional setting. The framework is demonstrated on oblique-view LPR, yielding a systematic empirical comparison of five deep-learning restoration architectures (U-Net, U-Net Conditional, Restormer, Pix2Pix, SR3 diffusion) on a shared $[0^\circ, 89^\circ]^2$ grid. Under the conditions of the LPR benchmark, approximately 90–92% of the evaluated angle grid is recoverable, depending on whether recovery is measured by the best single model or by the union of restoration arms, and the benchmark recoverability boundary occurs beyond roughly $80^\circ$ in both axes simultaneously, lateral rotations are consistently harder than elevational ones, and image-quality metrics track OCR accuracy closely enough to serve as a training-time surrogate. The instances of the framework beyond LPR are sketched in Section 6.2.

Relative to standard corruption-robustness benchmarks, the contribution is not another averaged robustness score over a finite list of corruptions. Instead, the framework estimates an operating boundary in a continuous acquisition-parameter space and reports both the size of the nominally recoverable envelope and the consistency of success inside that envelope.

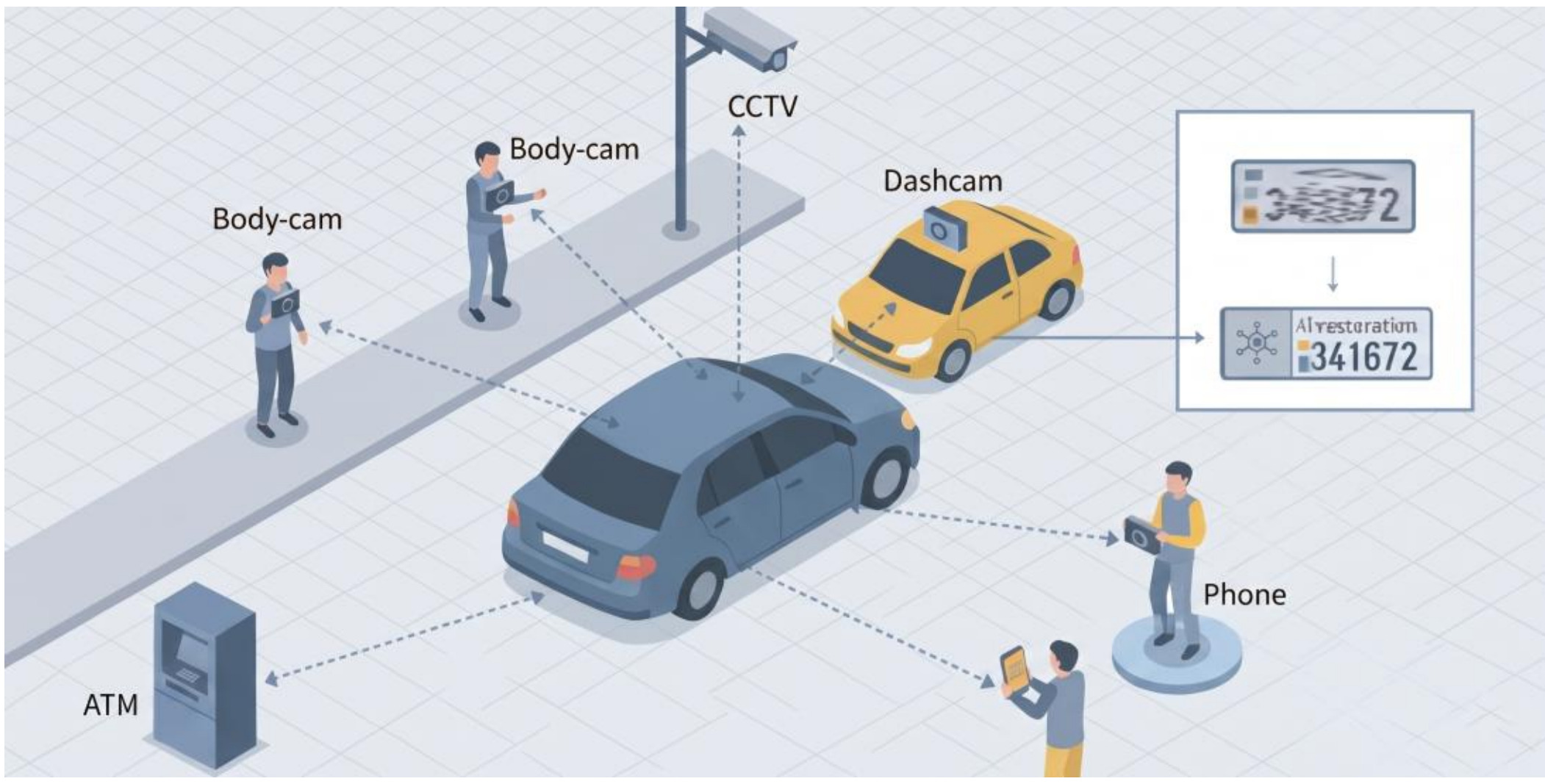

**Figure 1.** Opportunistic sensing in a smart city environment. Multiple imaging sensors deployed for unrelated purposes, ATM machine cameras, body-worn cameras, vehicle dashcams, pole-mounted CCTV, and handheld smartphones, incidentally capture a passing vehicle at diverse, uncontrolled viewing angles. This paper asks at which of these angles deep-learning image restoration can still recover a readable license plate from such non-purpose-built captures.

## 2. Related Work

### 2.1. Opportunistic and Repurposed Urban Sensing

Opportunistic sensing, in which sensors deployed for one purpose are secondarily exploited for another, has been systematically studied in mobile and pervasive computing. Burke et al. [9] introduced participatory sensing as a framework for community-scale data collection using everyday mobile devices; Campbell et al. [3] and Ganti et al. [4] extended this to opportunistic inference from always-on smartphone sensors. Lane et al. [2] demonstrated that commodity accelerometers, microphones, and cameras embedded in mobile phones can serve as general-purpose environmental monitors when paired with appropriate inference algorithms. In urban systems research, related concepts include volunteer geographic information [10] and the broader data-driven city management paradigm [11, 5].

The diversity of camera platforms in urban environments substantially widens the scope of opportunistic sensing beyond fixed infrastructure, following the broader pattern established by Eriksson et al. [1] for vehicle-mounted sensing without dedicated sensing missions.

Urban computing [12] and smart city frameworks [13] emphasise algorithmic repurposing of existing infrastructure to avoid capital expenditure. Two adjacent literatures inform the present study. First, urban camera geometry: the classical calibration framework of Zhang [14] provides the mathematical foundation for estimating the oblique pose of arbitrary fixed cameras from minimal scene content, a prerequisite for relating any empirical recoverability map to a specific urban sensor in the field. Second, privacy in secondary-use imagery: Frome et al. [15] documented large-scale face and license-plate blurring in Google Street View, establishing that extracting identity information from opportunistic captures carries well-understood ethical constraints which any deployment of the methods studied here must respect. Recent multimedia-security work on video cryptosystem analysis further emphasizes that camera-derived streams should be evaluated as security-sensitive multimedia data, not merely as computer-vision inputs [16]. Recent deep learning advances make such repurposing technically viable: where classical computer vision required controlled illumination, known geometry, and near-frontal capture, modern restoration pipelines can recover structured information from imagery that would previously have been discarded. The present work contributes a first quantitative angular boundary for when this AI-enabled repurposing succeeds for LPR under a controlled synthetic pipeline.

### 2.2. License Plate Recognition and Its Assumptions

LPR is a mature technology in controlled conditions. Comprehensive surveys [6, 7] document pipelines covering plate detection, character segmentation, and OCR, with commercial systems achieving near-perfect accuracy when plates are captured within $\pm 30^\circ$ of the optical axis. Oblique-view LPR pipelines exist, but they follow a detection-then-deskew paradigm that assumes the plate is already roughly readable before processing: homography estimation corrects for known moderate angles but breaks down once the plate is severely foreshortened.

Laroca et al. [17] constructed a large-scale benchmark for end-to-end automatic LPR and demonstrated that recognition accuracy degrades substantially at oblique capture angles even for state-of-the-art YOLO-based detectors, confirming the practical gap addressed here. Their detection-first pipeline implicitly assumes digit contrast and horizontal separation remain sufficient for segmentation, an assumption that breaks down once azimuthal rotation exceeds approximately $45^\circ$. Laroca et al. [18] later expanded this line with RodoSol-ALPR, explicitly evaluating cross-dataset generalization and showing that LPR performance can drop sharply when the evaluation distribution differs from training. Silva and Jung [19] proposed WPOD-NET to rectify oblique plates via per-instance planar warping before recognition, a detect-then-deskew approach that works well up to moderate angles but still requires the plate to be roughly readable. Xu et al. [20] built an explicit perspective-rectification module into the LPR pipeline. A parallel line of work treats oblique LPR as a super-resolution problem: Zou et al. [21] target extremely low-resolution plates, and Hamdi et al. [22] combine image enhancement with super-resolution to improve downstream OCR. The present work takes a different angle from these restoration-oriented studies: instead of proposing another restoration architecture, we map *where in the angle space* restoration remains viable at all, treating recoverability as a measurable property of the sensing geometry.

To anchor these observations quantitatively, recent methods on CCPD report high recognition accuracy on the benchmark's dedicated oblique subsets while leaving a clear residual gap in the most adverse regime: LPTR-AFLNet [23] reaches 99.2% on CCPD-Tilt and 99.2% on CCPD-Rotate, but 90.4% on CCPD-Challenge, so the difficulty concentrates where viewpoint and capture quality are jointly extreme. The standard remedies are geometric rectification before recognition [19, 20] and image restoration or super-resolution before OCR, which raises recognition on degraded RodoSol plates from 12.7% to 74.2% under an at-least-five-of-seven-characters criterion [24]. Each of these studies reports a single accuracy per subset or a fixed before-and-after gain at one operating point; the present work instead maps recognizability as a continuous function of viewing angle, locating the recoverability boundary that those discrete subset scores only sample.

The real-data validation uses CCPD as a public LPR dataset with vertex annotations and diverse parking-scene captures [25], and the cross-layout validation uses the Brazilian legacy and Mercosur car-layout slices from RodoSol-ALPR [18]. The OCR back ends used for sensitivity analysis span a classical document-OCR engine, Tesseract [26], a recent general OCR toolkit, PaddleOCR [27], a plate-specialized recognizer, HyperLPR3 [28], and an additional general OCR implementation, EasyOCR [29]. These tools are treated as measurement back ends rather than as restoration models: changes in their exact-match rates quantify how the same restored crops are read under different recognition assumptions.

Evaluation of recognition under degradation has a long history in related domains. PSNR and SSIM [30] are the standard image-quality proxies, but their correlation with downstream task performance is contested: for OCR and face recognition, PSNR is a more reliable predictor than SSIM [31]. This relationship has not been established for LPR under oblique-angle distortion. Prior recoverability analyses in audio and face recognition use threshold curves and operating-characteristic measures over a degradation parameter, but no such framework has been applied to the two-dimensional $(\alpha, \beta)$ space of camera viewing angles; the present work provides such a characterization.

The proposed maps also differ from conventional corruption-robustness benchmarks. Robustness benchmarks typically evaluate a model on a fixed catalogue of corruption types and severity levels, then aggregate accuracy over that catalogue. Such averages are useful for model ranking but can hide whether unrecovered cases are concentrated near a deployable boundary or scattered inside an apparently safe regime. Recoverability maps instead treat acquisition geometry as a continuous operating space: boundary-AUC measures the area enclosed by the outer success envelope, while $F$ records whether that envelope contains interior unrecovered pockets. This separation is important for opportunistic sensing because a sensor installation is governed by physical pose, not by a random draw from a generic corruption suite.

### 2.3. Deep Learning Image Restoration

The U-Net architecture [32], with its skip connections between encoder and decoder, remains a strong baseline for paired image-to-image tasks. The foundational U-Net architecture uses skip connections to preserve spatial detail through successive downsampling and upsampling stages. Oktay et al. [33] extended it with soft attention gates (Attention U-Net) that selectively emphasise encoder features relevant to the decoder prediction; a natural fit for localised digit regions on a plate. Transformer-based alternatives such as Restormer [34] and Uformer [35] extend global context modelling to restoration, achieving state-of-the-art results on denoising and deblurring benchmarks.

On the generative side, Goodfellow et al. [36] introduced the generative adversarial network (GAN) framework; Pix2Pix [37] adapted this to paired image translation via a patch discriminator. ESRGAN [38] demonstrated that perceptual and adversarial losses recover fine texture detail in super-resolution, at the cost of potential hallucination artefacts analogous to those we observe in Diffusion-SR3. Saharia et al. [39] proposed SR3, conditioning a denoising diffusion probabilistic model on a low-quality input for iterative refinement; we adapt this architecture directly for oblique-plate restoration. None of these works has been benchmarked on the systematic oblique-angle regime relevant to urban camera repurposing.

### 2.4. Low-Discrepancy Sampling for Benchmark Design

Monte Carlo methods for sampling multi-dimensional parameter spaces suffer from clustering and gaps at moderate sample sizes. Sobol sequences, a class of quasi-Monte Carlo methods [40, 41], provide provably better coverage of the unit hypercube than pseudo-random sampling: the discrepancy of an $N$-point Sobol set grows as $O((\log N)^d / N)$ compared to $O(N^{-1/2})$ for random sampling in dimension $d$. Applying Sobol sequences to the angle space ensures that the training dataset covers extreme and moderate angles uniformly, which is essential for learning a restoration model that generalizes across the full deployment range of a fixed urban sensor.

## 3. Problem Formulation

**Notation.** The following symbols are used consistently throughout the paper:

- $\alpha, \beta \in [-90°, 90°]$: primary variables, the azimuthal (lateral) and elevational (vertical) viewing angles of the plate relative to the camera optical axis.
- $x_0$, $x_d$, $\hat{x}_0$: clean plate image, observed distorted image, and restored plate image produced by the network.
- $\boldsymbol{R}(\alpha, \beta)$: homogeneous-coordinate rotation matrix; $\Pi$: perspective projection onto the image plane.
- $D_{\text{edge}}, D_{\text{jitter}}, D_{\text{blur}}, D_{\text{JPEG}}$: the four degradation operators (edge blending, colour jitter, Gaussian blur, JPEG compression) defined in Eq. 1.
- $r_T(\alpha, \beta) \in \{0,1\}$: binary recoverability indicator at OCR threshold $T$.
- $B_T$: recoverability boundary curve; AUC and $F$: coverage and reliability metrics defined in Eqs. 5 and 7.
- $f_\theta$: restoration network with parameters $\theta$.
- $\bar{\alpha}_t$ and $\beta_i$ (appearing only in Eq. 8): diffusion-model schedule coefficients; these are distinct from the viewing angles $\alpha, \beta$ and are used only in the SR3 formulation.

Consider a fixed camera at position $\boldsymbol{p}$ in an urban environment. A vehicle passes within the camera field of view, and its license plate undergoes a rotation described by two angles: $\alpha \in [-90°, 90°]$ (azimuthal, lateral tilt relative to the camera optical axis) and $\beta \in [-90°, 90°]$ (elevational, vertical tilt). The 3D rotation of the plate normal vector is modelled as successive rotations $\boldsymbol{R}(\alpha, \beta)$ applied in homogeneous coordinates, followed by a perspective projection $\Pi$ onto the image plane:

$$x_d = D_{\text{JPEG}} \circ D_{\text{blur}} \circ D_{\text{jitter}} \circ D_{\text{edge}} \circ \Pi(\boldsymbol{R}(\alpha, \beta)\, x_0) \quad (1)$$

where $x_0$ is the clean plate image, $\boldsymbol{R}(\alpha, \beta)\, x_0$ applies the planar rotation in homogeneous coordinates, $\Pi$ is the perspective projection, and the four degradation stages are applied in order: $D_{\text{edge}}$ performs alpha-mask edge blending with a signed-distance field, $D_{\text{jitter}}$ applies brightness/contrast/saturation jitter, $D_{\text{blur}}$ applies isotropic Gaussian blur with $\sigma \in [0.5,1.5]$, and $D_{\text{JPEG}}$ compresses the image at quality $q \in [55,85]$. The composition yields the observed distorted image $x_d$. Following the construction described in Section 4.1, $x_d$ is subsequently de-warped by the inverse homography $R(\alpha, \beta)^{-1}$ and resized to $256 \times 64$ before being passed to the restoration network; this realigned input is what each $f_\theta$ learns to correct. Thus the angular boundary reported in this paper is the boundary *after* inverse-homography compensation and reflects what information remains after resampling, not the raw oblique capture. The goal is to learn a restoration function $\hat{x}_0 = f_\theta(x_d)$ such that an OCR engine applied to $\hat{x}_0$ correctly reads the plate number.

Although the remainder of this paper specialises to LPR, the recoverability formalism introduced below is task-agnostic: given any parameterised degradation space $\mathcal{P}$ and any binary utility function $u: \mathcal{P} \to \{0,1\}$ indicating successful inference, the boundary-AUC and reliability score $F$ defined in Eqs. 5 and 7 carry over verbatim. In this paper $\mathcal{P} = [0°, 89°]^2$ is the oblique-angle grid and $u$ is thresholded plate-level OCR accuracy; other opportunistic-sensing tasks would instantiate $\mathcal{P}$ and $u$ differently but use the same machinery. Accordingly, the reported AUC and $F$ values should be read as task-conditioned system recoverability under the specified degradation model, OCR engine, and utility threshold, not as model-intrinsic performance independent of the sensing task.

Let $\text{OCR}_{\text{plate}}(\alpha, \beta) \in [0,1]$ denote the mean plate-level recognition accuracy (fraction of digits correctly identified) over a set of test images at angle pair $(\alpha, \beta)$. We define the binary *recoverability indicator* at threshold $T$:

$$r_T(\alpha, \beta) = \mathbf{1}\left[\text{OCR}_{\text{plate}}(\alpha, \beta) \geq T\right] \quad (2)$$

We set $T = 0.9$, requiring at least 9 out of 10 plates at each angle pair to be fully recognized. The *recoverability boundary* $B_T$ is the outer envelope of the region where $r_T = 1$. Formally, define (using the convention $\max\varnothing = -1$ so a fully unrecoverable slice is distinguishable from a boundary at $\beta = 0$):

$$\beta_{\max}(\alpha) = \max(\{-1\} \cup \{\beta \in [0,89] \mid r_T(\alpha, \beta) = 1\}) \quad (3)$$

$$\alpha_{\max}(\beta) = \max(\{-1\} \cup \{\alpha \in [0,89] \mid r_T(\alpha, \beta) = 1\}) \quad (4)$$

The boundary is $B_T = C_1 \cup C_2$ where $C_1 = \{(\alpha, \beta_{\max}(\alpha))\}$ and $C_2 = \{(\alpha_{\max}(\beta), \beta)\}$ for $\alpha, \beta \in [0,89]$. The *boundary-AUC* normalises the enclosed area by the full grid:

$$\mathrm{AUC} = \frac{1}{89^2} \cdot \frac{1}{2} \left( \int_0^{89} \beta_{\max}(\alpha)\, d\alpha + \int_0^{89} \alpha_{\max}(\beta)\, d\beta \right) \quad (5)$$

Boundary-AUC measures the area enclosed by the outer envelope of the recoverable region and therefore quantifies its *coverage*; by construction it does not penalise non-convex structure, for instance unrecovered holes inside the envelope, which can inflate AUC relative to the actually recoverable area. To capture the *consistency* of recoverability inside $B_T$, we measure how deeply unrecovered cases penetrate the recoverable region, a quantity conceptually related to erosion-style measures in mathematical morphology [42]. For each unrecovered angle pair $(\alpha_h, \beta_h)$ inside $B_T$, compute its minimum Euclidean distance to the boundary:

$$d_h = \min_{(\alpha_b, \beta_b) \in B_T} \sqrt{(\alpha_h - \alpha_b)^2 + (\beta_h - \beta_b)^2} \quad (6)$$

The *reliability score* $F$ is the root-mean-square of these distances, normalised by the enclosed area $E$:

$$F = \sqrt{\frac{1}{E} \sum_{h=1}^{N} d_h^2} \quad (7)$$

$F = 0$ denotes no interior unrecovered pockets; larger values of $F$ indicate more frequent or deeper unrecovered pockets. Taken together, boundary-AUC quantifies the extent of the viewing-angle space over which recovery succeeds, while $F$ measures the frequency and depth of unrecovered pockets within the nominally recoverable region. A model with high AUC and low $F$ covers a wide area uniformly; a model with high AUC but also high $F$ covers a wide area but behaves inconsistently; a model with low AUC and low $F$ covers less but has a more predictable boundary. OCR is used as the ground truth for recoverability because it measures the downstream utility of the restored plate. PSNR and SSIM are evaluated subsequently as candidate training-time surrogates for OCR (Section 5.5).

An intuitive way to read the two metrics is to imagine two sensors whose recoverable envelopes cover the same area. If Model A succeeds everywhere inside that envelope except at the outer boundary, it has high AUC and $F \approx 0$. If Model B reaches the same outer boundary but has isolated unrecovered holes deep inside the nominally safe region, its AUC can be identical while $F$ is larger. For deployment, AUC therefore answers "how far can the method reach?", whereas $F$ answers "how much can we trust the interior of that reach?". A high-AUC, high-$F$ model is useful for exploratory recovery but riskier for automated identity decisions than a high-AUC, low-$F$ model.

## 4. Methodology

The evaluation pipeline has six stages. First, random clean plate strings are rendered at a fixed canonical resolution. Second, Scrambled Sobol sequences sample the azimuth-elevation space so that moderate and extreme poses are both covered. Third, each clean plate is projected through the synthetic camera model, perturbed by blur, colour jitter, and JPEG compression, then inverse-warped back to the canonical plate frame. Fourth, each restoration model maps this degraded input to a same-resolution restored plate. Fifth, Tesseract OCR [26] converts the restored plate into a digit string and the result is compared with ground truth. Sixth, plate-level OCR success values over the full angle grid are converted into a recoverability map, boundary-AUC, and reliability score $F$. This workflow simulates opportunistic sensing by varying the camera-plate pose while holding the downstream recognition task and restoration protocol fixed. Figure 2 below summarises the recognition pipeline and the two recognition routes it compares (with and without restoration); the full per-stage procedure (render, Sobol-sample the pose, warp to oblique, degrade, dewarp, restore, read, score) is given in Section 4.1 and Eqs. 1, 5, and 7, so a single restoration model can be evaluated end to end from this description.

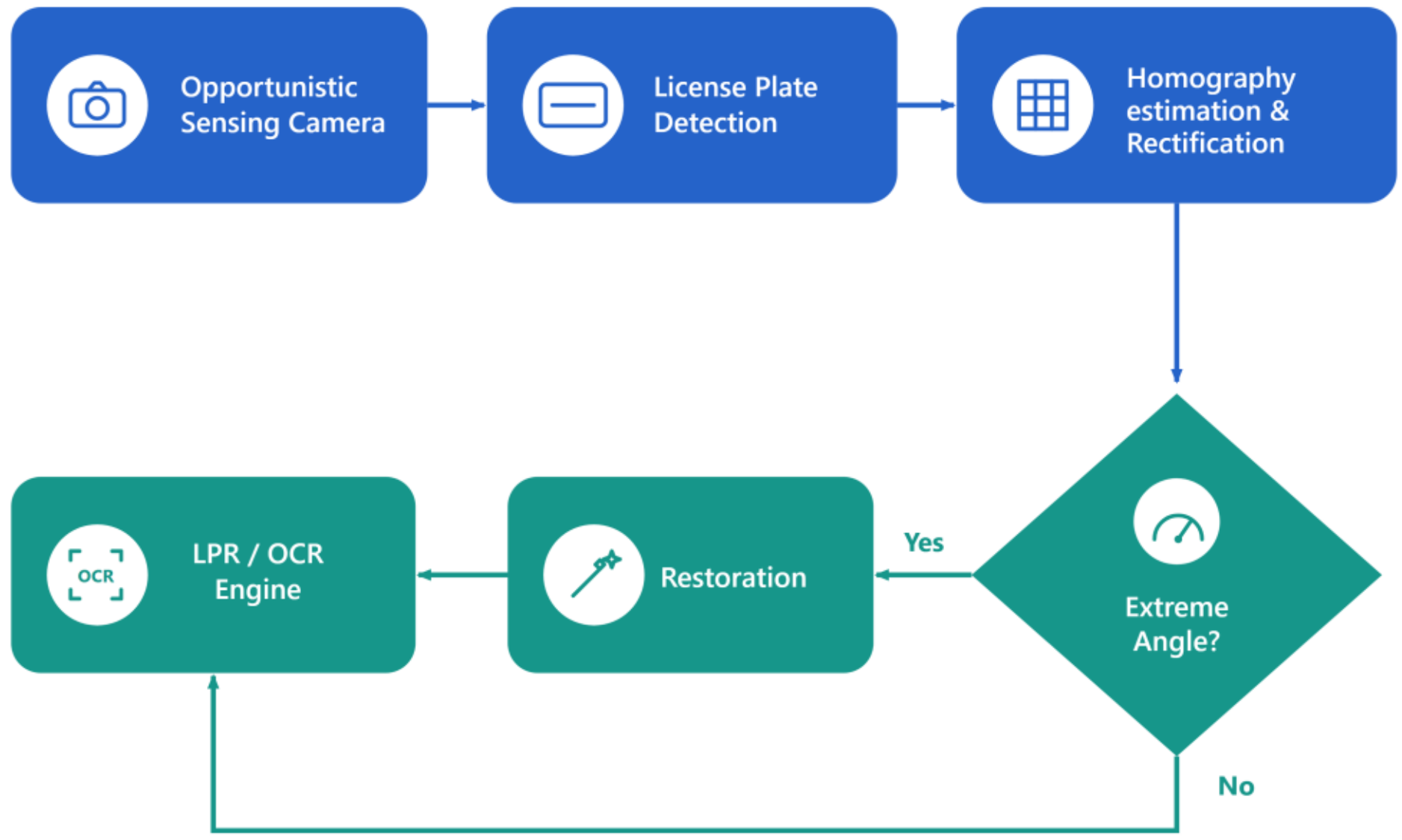

**Figure 2. Recognition pipeline.** An extreme-angle decision routes the rectified plate through a restoration step before the LPR/OCR engine, or straight to the engine when the capture is near-frontal. Sweeping the viewing angle $(\alpha, \beta)$ over the full grid turns the per-pose recognition outcomes into the recoverability map, boundary-AUC, and reliability score $F$ (Section 4.1; Eqs. 1, 5, 7).

### *4.1. Synthetic Dataset Construction*

We generate clean license plate images by rendering random 6-digit numeric strings on a yellow background using a single standard vehicle licence-plate font and dimensions, downscaled to $256 \times 64$ pixels (Figure 4). Each plate is then distorted through a four-stage pipeline:

- **Geometric warp:** homogeneous-coordinate rotation $\boldsymbol{R}(\alpha, \beta)$ followed by perspective projection $\Pi$ to produce a foreshortened plate (Eq. 1).
- **Edge blending:** a soft alpha mask computed via a signed-distance field and logistic smoothing suppresses boundary artefacts introduced by the projection.
- **Camera artefacts:** colour jitter (brightness, contrast, saturation), Gaussian blur ($\sigma \in [0.5, 1.5]$), and JPEG compression (quality $\in [55, 85]$) simulate sensor noise and compression.
- **De-warp and resize:** the distorted image is de-warped with the inverse homography and downscaled to $256 \times 64$ pixels, matching the clean image resolution.

Angle pairs $(\alpha, \beta)$ are drawn from a parametric probability density function (PDF) over $[-90°, 90°]^2$ using Scrambled Sobol sequences [41]. This provides near-uniform coverage of the angle space with far fewer samples than random sampling. We define two PDF variants: a *Standard* variant (DS-S) with moderate emphasis on extreme angles, and an *Extreme* variant (DS-E) with a heavier tail that concentrates more samples near oblique angles. Two datasets are constructed (Table 1); Figure 3 visualises their angle-density surfaces. All datasets use an 80/10/10 train/validation/test split. The grid deliberately covers the extreme oblique regime that real plate corpora rarely contain: across CCPD and the two RodoSol layouts no real plate reaches $60°$ on either axis (Appendix A.0, Figure A0).

**Table 1.** Dataset specifications. All datasets share the same distortion pipeline and $256 \times 64$ pixel resolution.

| Dataset | Total pairs | Angle PDF variant | Train | Validation | Test |
|---|---|---|---|---|---|
| Standard (DS-S) | 10,240 | Moderate emphasis on extreme angles | 8,192 | 1,024 | 1,024 |
| Extreme (DS-E) | 10,240 | Strong emphasis on extreme angles | 8,192 | 1,024 | 1,024 |

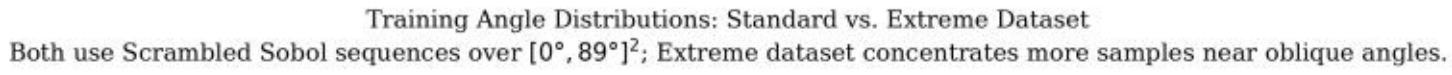

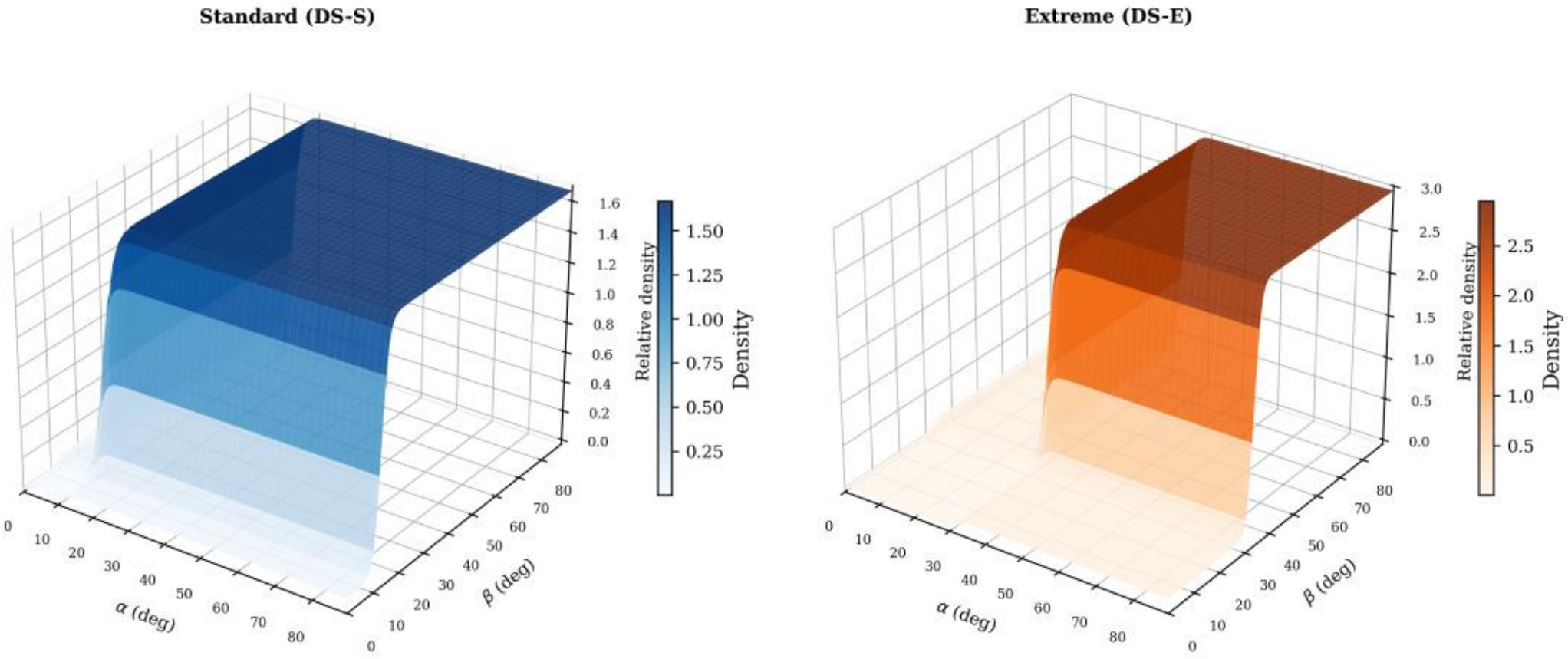

**Figure 3.** Angle-pair sampling density for the two training datasets, plotted as 3D surfaces over the $[0°, 89°]^2$ grid. Both datasets use Scrambled Sobol sequences for near-uniform base coverage. The Standard dataset (left, DS-S) applies a moderate logistic-shaped emphasis on large angles; the Extreme dataset (right, DS-E) concentrates substantially more mass near the oblique-angle corners, providing harder training examples and a more stringent test of each architecture's generalisation to extreme distortion.

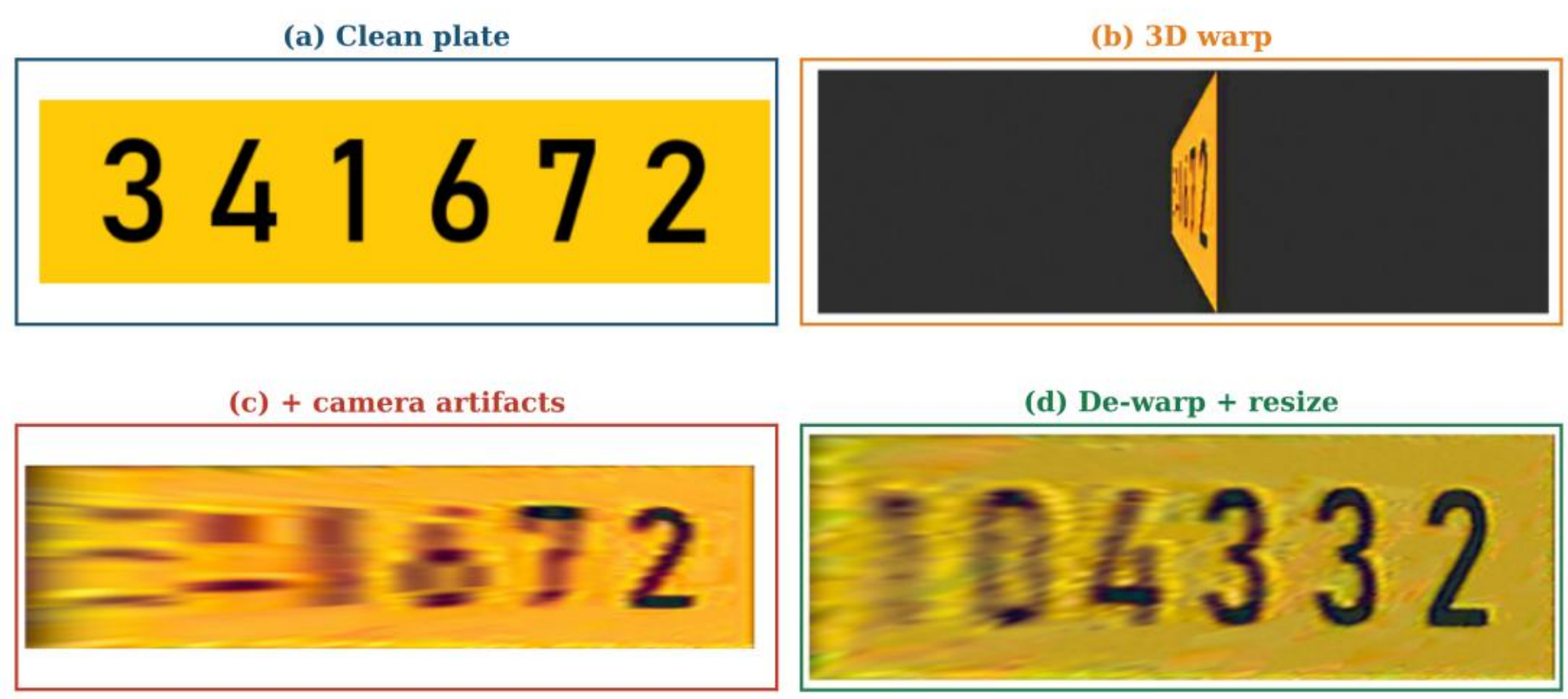

**Figure 4.** Synthetic training-data construction, shown as four stages. (a) A clean plate is rendered. (b) A 3D rotation $R(\alpha, \beta)$ and perspective projection $\Pi$ foreshorten it, where $\alpha$ is the azimuthal (lateral) rotation about the vertical axis and $\beta$ the elevational rotation about the horizontal axis. (c) Realistic camera artifacts (blur, colour jitter, JPEG compression) are added. (d) The plate is de-warped with $R^{-1}$ and resized to $256 \times 64$. Each restoration network learns to invert this pipeline.

*4.2. Restoration Architectures*

Five architectures spanning discriminative and generative paradigms are implemented and compared (Table 2). All models take the $256 \times 64$ distorted image as input and produce a same-resolution restored image.

The **U-Net baseline** uses a standard encoder-decoder with skip connections and an L1 + SSIM loss. The **U-Net Conditional** extends this with FiLM (Feature-wise Linear Modulation) layers: a lightweight encoder infers a distortion-pattern embedding from the input crop and injects it into each stage of the restorer, conditioning recovery on the inferred degradation. **Restormer** [34] replaces convolutional blocks with multi-Dconv head transposed attention (MDTA) and gated feed-forward networks, enabling long-range spatial modelling relevant to correcting directional foreshortening.

**GAN-Pix2Pix** [37] adds a patch discriminator to the U-Net generator, using a combined adversarial and L1 loss. **Diffusion-SR3** adapts the SR3 super-resolution diffusion framework [43] by conditioning the denoising network on the distorted plate. We use velocity prediction in place of noise prediction to stabilise training in the low-SNR uniform-colour regions of the plate:

$$v_t = \sqrt{\bar{\alpha}_t}\,\epsilon - \sqrt{1 - \bar{\alpha}_t}\,x_0 \tag{8}$$

where $\bar{\alpha}_t = \prod_{i=1}^{t}(1-\beta_i)$ is the cumulative signal coefficient and $\epsilon \sim \mathcal{N}(0, I)$. At inference, deterministic DDIM sampling [44] with 20 steps is used.

**Table 2.** Restoration architecture overview. Parameters reported for the specific configuration used in this study; all models accept $256 \times 64$ RGB input and produce a same-resolution restored output. Optimiser, base learning rate, and epoch count are the settings used in the reported experiments.

| Model | Key feature | Loss | Params(M) | Training Settings |
|---|---|---|---|---|
| U-Net (baseline)<br>*Discriminative* | Skip connections, encoder-decoder | L1 + SSIM | ≈ 7.8 | AdamW<br>$1\text{x}10^{-4}$, 120 epochs |
| U-Net Conditional<br>*Discriminative* | FiLM from inferred distortion embedding | L1 + SSIM | ≈ 8.6 | AdamW<br>$1\text{x}10^{-4}$, 120 epochs |
| Restormer<br>*Discriminative* | MDTA transformer, global context | L1 | ≈ 4.1 | AdamW<br>$3\text{x}10^{-4}$, 150 epochs |
| GAN-Pix2Pix<br>*Generative* | Patch discriminator, adversarial | Adversarial + L1 | ≈ 16.8 | Adam<br>$2\text{x}10^{-4}$, 200 epochs |
| Diffusion-SR3<br>*Generative* | Velocity prediction, DDIM (20 steps) | Velocity MSE | ≈ 12.9 | Adam + warm-up<br>$2\text{x}10^{-4}$, 600k steps |

### 4.3. Training Protocol

All models are trained on a single NVIDIA RTX 3090 (24 GB) GPU. Each model is trained independently on both datasets (Standard and Extreme), yielding ten model-dataset pairs. The per-model configuration is summarised in Table 2. Batch size is 32 for the three discriminative models, 16 for GAN-Pix2Pix (generator plus discriminator), and 16 for Diffusion-SR3. Discriminative models (U-Net, U-Net Conditional, Restormer) use the AdamW optimiser with weight decay $10^{-2}$ and a cosine-annealing learning-rate schedule decaying to $10^{-6}$ over the full training horizon. GAN-Pix2Pix uses Adam ($\beta_1 = 0.5$, $\beta_2 = 0.999$) with a fixed learning rate and alternates generator and discriminator updates every step. Diffusion-SR3 uses Adam with a 5000-step linear warm-up followed by constant learning rate, and is trained for a fixed step budget to match the optimisation budget despite its multi-step denoising objective; inference uses deterministic DDIM sampling with 20 steps. Model selection uses the validation-split PSNR and SSIM as surrogates for OCR: the checkpoint achieving the highest PSNR on the held-out validation set is retained for the final full-grid evaluation. Data augmentation is limited to the angle-sampling distribution itself (Section 4.1); no photometric augmentation is applied beyond the camera artefacts baked into the construction pipeline. Typical wall-clock training time per model-dataset pair is 3-5 h for U-Net variants, 35-45 h for Restormer, 4-6 h for GAN-Pix2Pix, and 12-18 h for Diffusion-SR3.

### 4.4. Full-Grid Evaluation

A single shared evaluation set covers all integer angle pairs in $[0,89]^2$ (8,100 pairs). Sampling density is 2 images per pair where $\alpha \leq 60$ and $\beta \leq 60$ (easy region), and 10 images per pair in the complementary high-distortion region ($\alpha > 60$ or $\beta > 60$). Six metrics are recorded per pair: plate-level PSNR and SSIM, worst-digit PSNR and SSIM, digit-level OCR accuracy, and plate-level OCR accuracy. The latter is the primary metric.

PSNR is computed as:

$$\text{PSNR} = 10\log_{10}\left(\frac{\text{MAX}^2}{\text{MSE}}\right) \quad (9)$$

OCR is performed by Tesseract v4 [26] (LSTM mode, digit-only filter). Each restored image is converted to grayscale, contrast-normalised, upscaled 2x, and binarised. A multi-strategy fallback (Otsu, adaptive threshold, colour inversion) is applied per digit when primary recognition is insufficient. The final recognised string is compared to the ground truth to compute digit-level and plate-level accuracy.

The real-data validation of §5.6 reuses this controlled-warp construction on real plate content. Because public LP datasets contain no extreme-angle captures, we exploit the ground-truth plate-corner annotations of CCPD and RodoSol-ALPR: a real plate is rectified to the canonical frame via its annotated corners, the same parameterised oblique warp is applied to synthesize a chosen high-angle view, and the result is dewarped back, yielding a held-out pair at a controlled extreme angle. This bridges the synthetic recoverability map to real plate content without requiring real extreme-angle captures, which do not exist in these corpora.

Two restoration-training regimes are reported. The main-grid models of Section 4.3 are trained on the fully synthetic pipeline and supervised to reconstruct the clean synthetic plate, so they learn to invert the synthetic degradation toward a synthetic target. For the real-data validation, a separate restorer is trained per dataset on the real-derived pairs above, taking the degraded dewarped crop as input and the plate's own real rectified frontal content as the target, so it learns to restore real plate appearance rather than a synthetic one; evaluation then uses held-out plates disjoint from this training set. The per-architecture recipes (Table 2) are identical across both regimes, so the recognition lift reported in Section 5.6 is attributable to the training content, not to a change of model.

## 5. Results

### *5.1. Training Performance on Dataset Test Splits*

Table 3 reports PSNR, SSIM, normalised training time, and inference latency for each model-dataset pair. Restormer achieves the highest PSNR and SSIM on both datasets. Diffusion-SR3 records lower PSNR/SSIM values, particularly on SSIM, where the gap to the U-Net baseline is 3.9–4.1 percentage points.

**Table 3.** Test-split PSNR (dB), SSIM, normalised training time (U-Net = 1.00), and inference latency (ms) per model and dataset. Best values per dataset shown in bold. Latency measured on a single NVIDIA RTX 3090 GPU (24 GB) using 256 × 64 RGB inputs, batch size 1, deterministic inference mode, and the same preprocessing pipeline used for the full-grid evaluation; reported values are per-image restoration latency and exclude OCR.

| **Model** | **Standard (DS-S)** | | **Extreme (DS-E)** | | **Train time (norm.)** | **Latency (ms)** |
|---|---|---|---|---|---|---|
| | **PSNR** | **SSIM** | **PSNR** | **SSIM** | | |
| U-Net (baseline) | 23.66 | 0.9705 | 20.96 | 0.9464 | 1.00 | 11.75 |
| U-Net Conditional | 24.18 | 0.9743 | 21.35 | 0.9508 | 1.19 | 7.50 |
| Restormer | 24.71 | 0.9762 | 21.67 | 0.9563 | 14.87 | 14.01 |
| GAN-Pix2Pix | 23.21 | 0.9672 | 19.97 | 0.9177 | 1.23 | 7.34 |
| Diffusion-SR3 | 21.74 | 0.9315 | 19.34 | 0.9052 | 4.69 | 21.81 |

On the Standard dataset, Restormer outperforms the U-Net baseline by 4.5% PSNR and 0.6% SSIM. U-Net Conditional improves by 2.2% PSNR at a training cost only 1.19x that of U-Net. GAN-Pix2Pix records a 2.0% lower PSNR value, and Diffusion-SR3 records 8.0% lower PSNR and 4.0% lower SSIM values. On the Extreme dataset the ranking is preserved, though absolute PSNR drops by ~3–4 dB for all models, reflecting the harder angle distribution (Table 3).

Efficiency results show that U-Net Conditional offers a favorable accuracy-latency tradeoff at 7.50 ms inference. Restormer requires 14.87x the training compute of U-Net and has 14 ms latency. Diffusion-SR3 is the most expensive to deploy at 21.81 ms due to its 20-step DDIM sampling loop.

### *5.2. Spatial Recognition Patterns*

Evaluating plate-level OCR accuracy over the full $[0°, 89°]^2$ angle grid reveals several spatial patterns that are consistent across all five models and both datasets. Recognition is near-perfect (OCR $\approx$ 1.0) when both $\alpha$ and $\beta$ are below 70°; this region covers the majority of the grid and is the basis for the high boundary-AUC values reported in Table 4. Recovery reaches the benchmark boundary once both angles exceed 80°, producing the small double-extreme unrecoverable corner quantified in Section 5.3.

Critically, the boundary is not symmetric. Along the extreme-$\beta$ edge, accuracy remains high for $\alpha$ up to $\approx 60°$ before declining. Along the extreme-$\alpha$ edge, accuracy drops sooner, even at low $\beta$. This $\alpha$-versus-$\beta$ asymmetry is consistent across both datasets and all five models, indicating that it is a property of the geometric distortion shared across architectures. A plate viewed from the side (high $\alpha$) loses digit separability faster than a plate viewed from above (high $\beta$) because lateral foreshortening compresses the horizontal extent where digit strokes reside, while vertical foreshortening still leaves digit contrast largely intact. Figure 6 (qualitative comparison) shows three regimes of increasing difficulty: at $(\alpha = 60°, \beta = 60°)$ every model recovers the plate, at $(\alpha = 75°, \beta = 73°)$ the discriminative restorers stay legible while the generative models begin to hallucinate digits, and $(\alpha = 78°, \beta = 78°)$ lies outside the reliable OCR region for every model.

*5.3. Recoverability Boundaries and Maximal Coverage*

Taking the pointwise maximum over all models and datasets yields the *maximal recoverability boundary*, which encloses a boundary-AUC of 0.934. In direct grid-count terms, approximately 90–92% of the $[0°, 89°]^2$ angle grid can be recovered at $\geq 90\%$ plate OCR accuracy, depending on whether recovery is measured by the best single model or by the union of restoration arms. The remaining double-extreme region (the upper-right corner, where both $\alpha$ and $\beta$ exceed $\sim 80°$) remains outside the recoverable envelope under the evaluated synthetic benchmark, restoration models, and Tesseract-based OCR pipeline [26]: the signal energy in the severely foreshortened plate image is too low for these evaluated methods to extract a recognisable digit sequence. Across all ten model-dataset pairs the per-model boundaries cluster tightly (see Table 4), confirming that the recoverable region is primarily shaped, under this benchmark, by the geometry of foreshortening, with architecture and training distribution affecting secondary properties such as efficiency and interior consistency. The $\alpha$-axis contracts faster than the $\beta$-axis, confirming the lateral rotation asymmetry.

*5.4. Recoverability-Reliability Tradeoff*

Table 4 summarises the boundary-AUC and reliability score $F$ for all model-dataset pairs. Discriminative models cluster in the AUC range $[0.920, 0.927]$ with $F \in [0.094, 0.207]$, indicating both wide coverage and consistent interior performance. Generative models fall 1–3% lower in AUC: GAN-Pix2Pix achieves $F \approx 0.17$, while Diffusion-SR3 shows highly variable reliability, from $F = 0.57$ on Standard to $F = 1.12$ on Extreme. This instability reflects the tendency of diffusion models to hallucinate plausible but incorrect digit sequences when the conditional input carries insufficient signal.

Restormer has the highest AUC point estimate on both datasets (0.926 and 0.927), while U-Net Conditional provides a favorable accuracy-cost tradeoff and the lowest $F$ score on the Extreme dataset (0.115). On Extreme, Restormer's $F$ rises to 0.207; its coverage is therefore robust to training-distribution shift, while U-Net Conditional offers the more favourable reliability-efficiency profile. Comparing Standard and Extreme datasets, AUC values are nearly identical ($\Delta < 0.002$), indicating that training distribution shift does not substantially alter the recoverable boundary itself, only its interior reliability.

**Table 4.** Boundary-AUC (with 95% confidence intervals from a stratified cluster bootstrap over grid cells) and reliability score $F$ per model and dataset on the full $[0°, 89°]^2$ evaluation grid ($T = 0.9$). Higher AUC and lower $F$ are better; best AUC per dataset in bold. The AUC intervals are tight and overlap among the discriminative models, while the reliability score $F$ is far more variable: Diffusion-SR3's $F$ interval spans $[0.85, 1.40]$ on DS-E, consistent with its hallucination-driven interior unreliability.

| Model | Standard (DS-S) | | Extreme (DS-E) | |
|---|---|---|---|---|
| | AUC | F (deg) | AUC | F (deg) |
| U-Net (baseline) | 0.920 [.917, .923] | 0.102 | 0.920 [.917, .923] | 0.146 |
| U-Net Conditional | 0.922 [.919, .925] | 0.144 | 0.923 [.920, .926] | 0.115 |
| Restormer | 0.926 [.923, .928] | 0.094 | 0.927 [.924, .929] | 0.207 |
| GAN-Pix2Pix | 0.915 [.912, .918] | 0.168 | 0.905 [.901, .908] | 0.163 |
| Diffusion-SR3 | 0.894 [.891, .897] | 0.569 | 0.892 [.888, .895] | 1.122 |

*5.5. Metric Proxy Validity*

For system developers, a key practical question is whether PSNR or SSIM can serve as a reliable surrogate for OCR accuracy during model training and validation, avoiding the computational cost of running OCR on every epoch checkpoint.

Both PSNR and SSIM track plate OCR accuracy closely enough to serve as a training-time surrogate (Figure 5). Per-character OCR rises with each metric and then saturates once the plate becomes legible, so the relationship is well described by a saturating curve rather than a straight line. The rank correlation with OCR is high for both metrics and marginally higher for SSIM: Pearson $r = 0.89$ (Restormer) and $0.91$ (Diffusion-SR3) for PSNR, versus $r = 0.98$ and $0.95$ for SSIM. SSIM shares the bounded $[0,1]$ range of OCR accuracy and saturates together with it, whereas PSNR is unbounded in decibels and keeps rising after OCR has saturated.

Either metric therefore supports OCR-free model selection. We report PSNR as the default training surrogate because it is loss-aligned with the L1/MSE objective, and SSIM as an equally valid, marginally tighter alternative. Both calibrations are benchmark-bound: their generality to other plate colours, font weights, national formats, and OCR back-ends should be verified before either is used as a deployment rule rather than a training convenience.

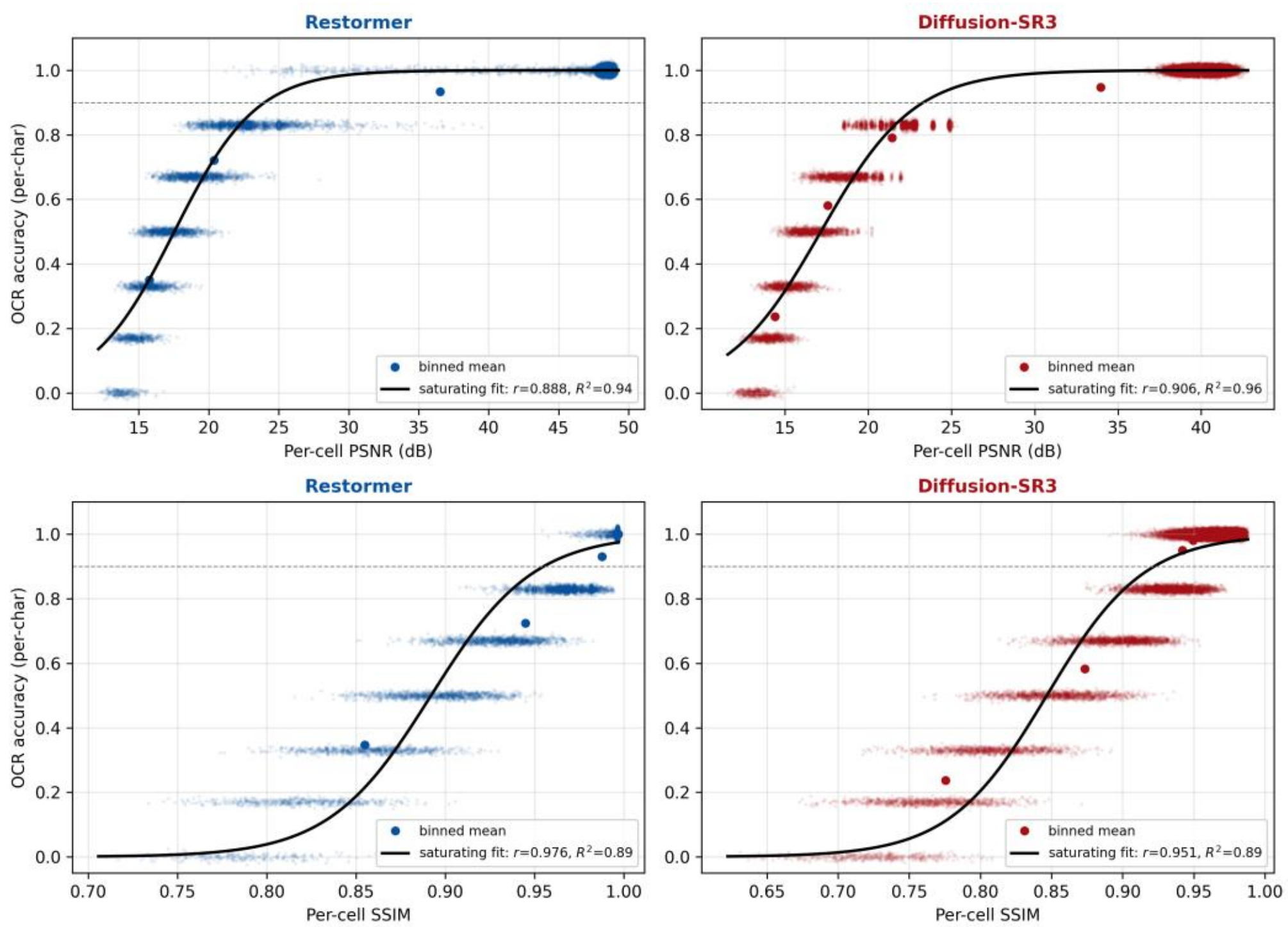

**Figure 5.** Image-quality metrics as OCR surrogates on the evaluation grid. Rows: per-cell PSNR (top) and SSIM (bottom); columns: Restormer (blue) and Diffusion-SR3 (red). Each panel shows the per-sample scatter, binned means, and a saturating (logistic) fit with Pearson $r$ and $R^2$. Both metrics track per-character OCR, which saturates as the plate becomes legible; SSIM has the marginally higher rank correlation because it is bounded and saturates together with OCR, while PSNR is unbounded in decibels. The dashed line marks the $T = 0.9$ OCR level.

**Figure 6.** Qualitative restoration at three angle regimes of increasing difficulty: good $(60°, 60°)$, partial $(75°, 73°)$, and outside the reliable band $(78°, 78°)$. Top row: the stored degraded input; second row: ground truth; below: each model. Discriminative restorers (U-Net variants, Restormer) stay legible through the partial case, while generative models (GAN-Pix2Pix, Diffusion-SR3) hallucinate plausible but incorrect digits under low signal; in the hardest case every model mis-recovers the leading digit.

### 5.6. Real-Data Validation

The recoverability boundary above is established on the controlled synthetic grid; we validate that restoration also improves recognition on real plates. Three real datasets, CCPD [25] and the Brazilian legacy and Mercosur layouts of RodoSol-ALPR [18], were selected because they provide ground-truth plate-corner annotations, which yield an exact rectifying homography: each real plate is dewarped to a canonical frontal view and re-warped to controlled oblique angles (the semi-synthetic construction of §4.4), supplying the extreme regime that no real capture samples (all three datasets top out below 60°). For every engine the *before* rate is that engine reading the degraded dewarped crop and the *after* rate is the same engine reading the restored crop: the dewarp restores the plate's frontal shape but not the detail lost to foreshortening and degradation, so each engine, including the general-purpose ones, still produces a (low) reading before restoration. Treating restoration as an extreme-angle module, a standard recognizer already reads near-frontal plates, we evaluate on held-out test plates excluded from restorer training. Each dataset is scored by its plate-specialised recognizer, which already reads the degraded plate on fair terms: HyperLPR3 for CCPD and Fast-Plate-OCR for the two RodoSol layouts. Restoration lifts whole-plate exact-match across all three datasets, most strongly in the extreme regime DS-E ($> 45^\circ$). Scored by these plate-specialised recognizers, the discriminative restorers lift exact-match by +15 to +38 points; the best restorer per dataset reaches 80.8% on CCPD, 86.1% on the Brazilian layout, and 96.5% on Mercosur (Table 5). The full per-architecture results are in Appendix A.1.5. General-purpose OCR (EasyOCR, Tesseract), reading the cropped character region, gains even more in absolute terms, up to +57 points at extreme angles where the degraded baseline is lowest (Appendix Table A1), with character accuracy (1 − CER) spanning 52 to 95%.

**Table 5.** Real-data recognition in the extreme regime (DS-E, worst-axis angle $> 45^\circ$), before and after restoration, on held-out test plates (excluded from restorer training), scored by each dataset's *plate-specialised* recognizer (HyperLPR3 for CCPD, Fast-Plate-OCR for the two RodoSol layouts), plus the strong general-purpose PaddleOCR on CCPD. The *same* recognizer reads both the degraded and the restored plate, so the degraded → restored change isolates the restorer with recognition held fixed; each reads the degraded plate on fair terms. Each row uses the best restoration architecture on that dataset (U-Net for CCPD, Restormer for RodoSol). Cells report whole-plate exact-match and character accuracy (1 − CER), degraded → restored.

| Dataset | Recognizer | Restorer | *n* | degraded → restored | |
|---|---|---|---|---|---|
| | | | | exact-match | 1 − CER |
| CCPD | HyperLPR3 | U-Net | 400 | 42.5 → **80.8** | 66.5 → 94.1 |
| CCPD | PaddleOCR | U-Net | 400 | 38.5 → **77.5** | 64.8 → 93.3 |
| Brazilian | Fast-Plate-OCR | Restormer | 941 | 63.2 → **86.1** | 80.7 → 96.8 |
| Brazilian | PaddleOCR | Restormer | 941 | 63.8 → **87.8** | 86.0 → 97.4 |
| Mercosur | Fast-Plate-OCR | Restormer | 941 | 66.0 → **96.5** | 83.0 → 99.3 |
| Mercosur | PaddleOCR | Restormer | 941 | 53.6 → **96.7** | 74.6 → 99.4 |

The architecture recommendation of §5.4, prefer discriminative restorers over generative ones, reproduces on real plates. Training the same five architectures on CCPD under their per-architecture recipes and measuring boundary-AUC over a dense $(\alpha, \beta)$ grid, the two discriminative restorers U-Net and Restormer expand the recoverable region (boundary-AUC 0.445 → 0.600 and 0.445 → 0.522), while the generative Pix2Pix GAN collapses it far below the degraded input (0.445 → 0.050) and Diffusion-SR3 is flat (0.453): the GAN hallucinates clean, confident, incorrect glyphs even on already-readable standard-angle plates. On real plates the discriminative advantage is thus visible in coverage itself, not only in interior reliability. The same ordering holds on the two RodoSol layouts under their plate-specialised recognizer (Fast-Plate-OCR): the discriminative restorers lead, Restormer best on both (86.1% Brazilian, 96.5% Mercosur), while the generative Diffusion-SR3 gains the least and on the Brazilian layout drops below the unrestored plate (53.6% versus 63.2%), making the generative identity-harm directly visible on real plates. Full per-engine sensitivity, per-angle-bin recoverability, training-regime and architecture ablations, and restoration examples are in Appendix A.1.

## 6. Discussion

### 6.1. Interpretation of the Observed Patterns

Three empirical patterns follow from the structure of the problem, not from implementation details, and a fourth concerns how the training-angle distribution affects interior consistency more than coverage. The $\alpha/\beta$ asymmetry reflects the information geometry of the plate itself: a license plate is a horizontal strip in which digit identity is encoded in the horizontal separation of character strokes more than in vertical extent. An elevational rotation $\beta$ compresses vertical extent but leaves the horizontal digit sequence intact, whereas a lateral rotation $\alpha$ compresses the axis along which the digits are encoded, collapsing adjacent characters into each other long before the plate becomes geometrically unrecognisable. This asymmetry is visible in every model and is therefore a property of the input signal, not an artifact of the networks.

The advantage of discriminative models reflects an alignment between training objective and task. U-Net, U-Net Conditional, and Restormer are trained end-to-end with paired (distorted, clean) samples and an L1 + SSIM objective, which is a direct regression of the distortion-to-character map that the networks must invert at inference time. GAN adversarial losses and diffusion denoising objectives instead prioritise perceptual plausibility, a criterion that competes with character fidelity: a plausible plate need not be the correct plate. Restormer and the U-Net variants therefore cluster near the top of Table 4 on both AUC and $F$. Diffusion-SR3's instability at extreme angles follows from the same tradeoff: when the conditional input carries low signal-to-noise the denoising network can sample a digit string consistent with the remaining evidence, producing plausible but incorrect plates. The reliability score $F$ captures this directly, rising to $1.122$ on DS-E, larger than the corresponding discriminative-model values. Figures 6 and A5 provide representative qualitative examples of this behaviour: under low-signal viewpoints, generative outputs can look cleaner than the input while no longer preserving the correct digit identity.

The training distribution shapes interior consistency more than coverage. Boundary-AUC is nearly identical across the Standard and Extreme datasets ($\Delta < 0.002$ for the discriminative models), indicating that the extent of the recoverable region is dominated, in this benchmark, by foreshortening and resampling geometry, with a secondary contribution from the training-angle PDF. The angle PDF does, however, shape $F$: the Extreme dataset destabilises Diffusion-SR3 ($F$ rises from $0.57$ to $1.12$) and produces small shifts in interior behaviour for the other models. This is the classical bias-variance tradeoff of data distribution design and implies that practitioners building real-data pipelines should prioritise angular diversity and corner coverage over raw sample count.

### *6.2. Beyond LPR: Other Instances of the Framework*

The recoverability-map framework introduced in Section 3 is deliberately task-agnostic, and the methodology demonstrated here on oblique-view LPR transfers to several other opportunistic-sensing problems of interest to the urban-systems community. Face recognition from pole-mounted or body-worn cameras under controlled pose and occlusion parameters fits the framework directly: the parameter space becomes (yaw, occlusion fraction) and the utility is a thresholded face-identification score. Scene-text reading from dashcam imagery under glare and motion-blur parameters is another natural instantiation. Gait recognition from non-frontal CCTV, where the parameter space is camera viewpoint and walking-direction offset, is a third. In each case the same two summary measures, boundary-AUC and the reliability score $F$, would yield a comparable quantitative deployment guideline. These instantiations define a natural extension path, with LPR serving here as a concrete first demonstration that the framework produces actionable numbers.

### *6.3. Implications for Smart City Deployment*

The approximately 90–92% recoverable-grid envelope provides a benchmark-specific first-order planning hypothesis for urban infrastructure, not a field-ready deployment guarantee. Under the assumptions of the synthetic benchmark, if the worst-case viewing angle for a fixed sensor installation is bounded by $\alpha < 80^{\circ}$ and $\beta < 80^{\circ}$, then software-only restoration *appears capable* of recovering legible plates without hardware modification; this envelope covers the majority of practical ATM-camera, body-worn-device, and streetlight-camera geometries, whereas sensors placed with extreme lateral tilt and extreme elevation simultaneously fall outside it. Translating this guideline into a deployment criterion requires controlled real-camera validation, calibration of the synthetic degradation model to the target camera class, and evaluation with the OCR engine used in the operational pipeline.

Among the evaluated architectures, discriminative models are the clear operational choice: they are easier to validate through explicit loss functions, more stable to train, and their PSNR improvements translate directly to OCR gains (Figure 5). The hallucination behaviour of Diffusion-SR3 under low signal requires additional identity-preservation safeguards, since the model produces a visually clean plate with plausible but incorrect digits, a confident incorrect reading that should be routed through independent confidence checks. Discriminative models, by contrast, produce blurred or ambiguous outputs under the same conditions, which such filters can reject. Among discriminative models, U-Net Conditional offers a favorable operational tradeoff: accuracy within 2% of Restormer, only 1.19x the training cost of the U-Net baseline, and the lowest inference latency among the discriminative restorers (7.50 ms), which is compatible with real-time processing of continuous fixed-sensor streams at over 100 frames per second on a single GPU.

The $\alpha/\beta$ asymmetry gives a practical corollary for camera placement: lateral mounting angle (which determines $\alpha$) is a more critical variable than vertical elevation (which determines $\beta$). A camera mounted very high on a pole retains LPR capability across a wide range of vehicle distances because the plate remains partially visible; a camera angled steeply to the side loses digit separability earlier. Urban planners seeking to enable opportunistic LPR from pole-mounted sensors should prioritise constraining the azimuthal angle over the elevational angle in sensor placement guidelines.

Two additional deployment use-cases follow directly from the recoverability map:

- **Forensic triage.** For retrospective case work on archived footage, the recoverability map supports fast pre-screening: captures whose estimated $(\alpha,\beta)$ fall in the upper-right corner ($\alpha,\beta > 80°$) can be flagged as unlikely to be recoverable under the evaluated method family, saving analyst time while leaving final evidentiary decisions to validated operational procedures.
- **Expected-gain estimation.** Given a sensor portfolio with known geometry distribution, overlaying that distribution on the recoverability map gives the *expected* fraction of captured vehicles that a restoration pipeline will render readable, a quantity directly useful for ROI and procurement decisions.

*6.4. Ethics, Privacy, and Security*

Opportunistic LPR is intrinsically sensitive because it turns imagery captured for one purpose into identity-bearing data for another. Any deployment of the methods studied here should be bounded by a lawful purpose, explicit retention limits, access control, audit logging, and human review of low-confidence or high-consequence outputs. The hallucination behaviour observed for generative restoration is especially relevant ethically: a visually plausible but incorrect plate can create a false identity association. For that reason, restoration outputs should not be treated as ground truth without OCR confidence checks, provenance records, and, where possible, independent corroborating evidence.

The multimedia security dimension is also material. Following recent work on cryptanalysis and strengthening of video cryptosystems [16], opportunistic camera streams and restored plate crops should be protected in transit and at rest, and intermediate restored images should be handled as sensitive derived data with retention and access controls. Security controls such as encryption, role-based access, tamper-evident logs, and deletion policies are outside the algorithmic contribution of this paper, but they are necessary conditions for responsible use in smart-city or forensic workflows.

*6.5. Scope Conditions and Validation Envelope*

The study has several bounded scope conditions that should be read alongside the headline results. The benchmark is fully synthetic: clean plates are rendered in a single standard vehicle-plate format at a fixed $256 \times 64$ resolution, distorted by a fixed homographic pipeline at constant focal length and plate-to-camera distance, and perturbed with Gaussian blur ($\sigma \in [0.5,1.5]$), colour jitter, and JPEG compression (quality $\in [55,85]$). Real captures introduce additional effects not modelled here, including scale variation with vehicle distance, motion blur, rolling-shutter skew, low-light noise, multi-font plate designs, dirt, and partial occlusion. Absolute recoverability thresholds reported in this paper, including the approximately 90–92% recoverable-grid envelope, therefore belong to the benchmark; the *structure* of the recoverability space (the location of the boundary, the $\alpha/\beta$ asymmetry, and the benchmark architecture comparison) is the transferable contribution.

The OCR back-end is Tesseract v4 [26] with standard preprocessing (grayscale, contrast normalisation, $2\times$ upscaling, binarisation, per-digit fallbacks). A stronger end-to-end plate-specific OCR (for example PaddleOCR [27] or a CRNN variant) would likely shift the absolute boundary numbers, though the relative ranking of architectures is expected to remain stable; this expectation should be verified experimentally before claiming OCR-independent recoverability boundaries. Appendix A adds real-data recognition validation, OCR-engine sensitivity checks, an identity-preservation analysis, a training-set-size convergence ablation, and confidence-interval and seed-stability audits. The main results table remains the primary synthetic benchmark summary, while the appendix defines the current validation envelope and the next extensions needed for deployment-transfer validation. The real-data validation is moderate-angle by nature: the evaluated corpora contain no plate beyond 60 degrees, so the extreme-angle recognition evidence (Appendix A.1.3, Table A1) is measured on a semi-synthetic construction that warps real plate content to extreme angles through the training pipeline, evaluated on held-out test plates excluded from restorer training.

## 7. Conclusion

This paper introduced *recoverability maps*, a task-agnostic framework for quantifying where in a parameterised degradation space an opportunistic-sensing task remains reliable. The framework builds on two summary measures adapted from classical operating-characteristic analysis: a boundary area-under-curve that captures coverage, and a reliability score $F$ that captures interior consistency. We demonstrated the framework on oblique-view license plate recognition as a concrete first instance.

Applied to LPR, the framework yields a sharp quantitative envelope. Approximately 90–92% of the $[0°, 89°]^2$ angle grid is recoverable at $\geq 90\%$ plate OCR accuracy within the evaluated synthetic benchmark, depending on whether recovery is measured by the best single model or by the union of restoration arms; beyond roughly $80°$ in both axes simultaneously, none of the evaluated architectures extracts a reliable plate number under the tested model and OCR pipeline. The most intellectually load-bearing finding, however, is structural: all five restoration architectures share the same $\alpha/\beta$ asymmetry, in which lateral rotation is consistently harder to correct than elevational rotation. This asymmetry is a property of the input signal, namely the horizontal information geometry of the plate, not of any particular network, and it carries directly into camera-placement guidance for urban infrastructure. Discriminative architectures (U-Net, U-Net Conditional, Restormer) consistently outperform generative alternatives (GAN-Pix2Pix, Diffusion-SR3) in coverage, consistency, and reliability; generative hallucinations under low signal create an identity-preservation risk for any identity-recognition application. Image-quality metrics (PSNR and SSIM) track OCR accuracy closely enough to serve as training-time surrogates within this benchmark, which materially simplifies the model-selection loop for the evaluated plate format. OCR-in-the-loop validation remains necessary when changing the OCR engine, plate design, or camera domain.

Within the benchmark, most oblique viewing angles yield recoverable plates, a small double-extreme corner of the angle grid does not, and these findings, together with the framework itself, transfer to a broader class of opportunistic-sensing problems. The absolute numbers belong to the benchmark; the structural findings and the framework are the transferable contribution.

A natural future-work direction follows from the observation, already raised in the Introduction, that public benchmarks do not sample the extreme parameter regimes in which opportunistic sensing actually operates. Closing this gap, for LPR and for the candidate tasks listed in Section 6.2, requires either synthetic benchmarks of the kind presented here (extended to additional tasks and to richer acquisition models) or bespoke controlled-capture datasets that systematically sweep the parameter space. Within LPR specifically, this study includes shared-sample multi-engine OCR sensitivity and a homography-binned real-camera high-angle validation protocol. Complementary next steps are an AUC/F/OCR seed matrix extending the completed primary U-Net seed audit and broader plate-format coverage.

## Author Contributions

Conceptualization, A.A. and Y.A.; methodology, I.A., O.B.A. and A.A.; software, I.A. and O.B.A.; validation, I.A., O.B.A. and A.A.; formal analysis, I.A., O.B.A. and A.A.; investigation, I.A. and O.B.A.; data curation, I.A. and O.B.A.; writing, original draft preparation, I.A., O.B.A. and A.A.; writing, review and editing, Y.A. and A.A.; visualization, I.A. and O.B.A.; supervision, Y.A. and A.A.; project administration, A.A. All authors have read and agreed to the published version of the manuscript.

## Funding

This research received no external funding.

## Institutional Review Board Statement

Not applicable. This study did not involve humans or animals; all license plate imagery used in the benchmark is synthetically generated with randomly sampled digit strings.

## Informed Consent Statement

Not applicable.

## Data and Code Availability

The companion repository for this paper is hosted at https://github.com/ApartsinProjects/OpportunisticSensingV2. The repository contains the paper HTML, all figures used in the manuscript, the figure-generation scripts required to reproduce the plots, the five restoration-architecture implementations (U-Net, U-Net Conditional, Restormer, Pix2Pix, SR3 diffusion), and notebooks covering the Scrambled Sobol angle sampling, the full-grid evaluation harness, and the Tesseract v4 OCR pipeline [26] described in Section 4. The synthetic datasets analysed in this study are not redistributed as static files; they are fully and deterministically reproducible by re-running the sampling notebook with the Sobol seeds and distortion-pipeline parameters documented in Section 4.1 and Table 1, which yields bit-identical training, validation, and test splits. Figure-level data files, generated figures, and supporting evaluation artifacts are included in the repository. The fifteen trained restoration checkpoints (the five architectures across CCPD and the Brazilian legacy and Mercosur RodoSol layouts) are archived on Zenodo at https://doi.org/10.5281/zenodo.20678637, so that all table values can be recomputed without corresponding-author mediation.

## Conflicts of Interest

The authors declare no conflict of interest.

## Abbreviations

The following abbreviations are used in this manuscript:

| | |
|---|---|
| AUC | Area under the curve |
| CCTV | Closed-circuit television |
| DDIM | Denoising Diffusion Implicit Models |
| DET | Detection Error Tradeoff |
| DS-E | Extreme dataset variant |
| DS-S | Standard dataset variant |
| GAN | Generative adversarial network |
| GPU | Graphics processing unit |
| JPEG | Joint Photographic Experts Group |
| LPR | License plate recognition |
| OCR | Optical character recognition |
| PDF | Probability density function |
| PSNR | Peak signal-to-noise ratio |
| RGB | Red-green-blue |
| ROI | Region of interest |
| SR3 | Super-Resolution via Repeated Refinement |
| SSIM | Structural similarity index measure |
| U-Net | U-shaped convolutional encoder-decoder network |

## Appendix A. Supplementary Experiments

This appendix reports supplementary experiments that extend the main synthetic recoverability study. Each experiment is presented with its protocol, metric, and result. The experiments address real-data validation, OCR-engine sensitivity, high-angle behavior, plate-style robustness, generative restoration artifacts, sampling convergence, uncertainty, seed stability, cross-layout validation, and degradation analysis.

*A.0. Exploratory data analysis: viewing-angle coverage of the real datasets*

The three real datasets evaluated in this appendix are dominated by near-frontal and moderate-angle captures. Decoding the worst-axis viewing angle of each plate (homography-decoded max(|yaw|, |pitch|) for CCPD, and proxy yaw/pitch for the RodoSol layouts) gives a median of 20° for CCPD ($n$ = 28,681, the complete local corpus decoded

from each image's filename-encoded plate corners), 12.5° for the Brazilian legacy layout ($n = 5000$), and 12.6° for Mercosur ($n = 5000$), with maxima of 59°, 38°, and 36° respectively. No plate in any of the three datasets reaches 60° on either axis (Figure A0), so the extreme oblique regime mapped by the controlled synthetic benchmark is essentially unsampled by these real corpora. All three datasets were selected precisely because they provide ground-truth plate-corner annotations (CCPD encodes the four plate vertices per image; RodoSol-ALPR labels the plate quadrilateral): the corners yield an exact rectifying homography that dewarps each real plate to a canonical frontal view, which the semi-synthetic construction (Appendix A.1) then re-warps to controlled extreme angles. The extreme regime absent from every real capture is therefore synthesized from each plate's own real content rather than approximated, so the high-angle recognition evidence below is geometrically faithful to each dataset's plates, while the recoverability boundary itself remains a controlled synthetic-benchmark result.

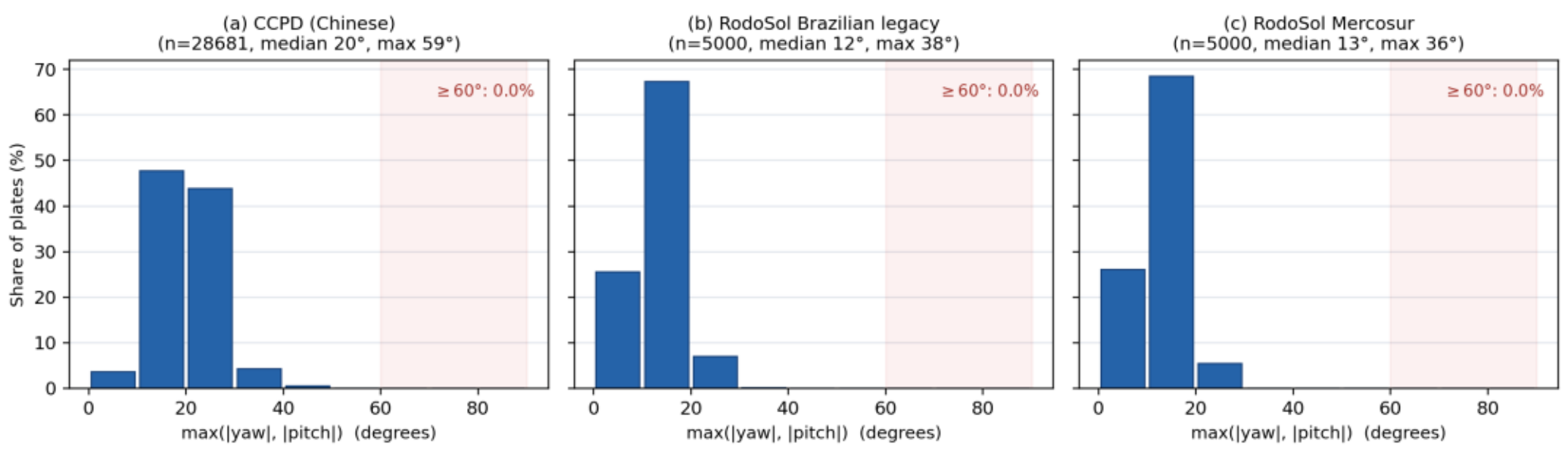

**Figure A0.** Distribution of the worst-axis viewing angle max(|yaw|, |pitch|) for the three evaluated real datasets: (a) CCPD, (b) RodoSol Brazilian legacy, and (c) RodoSol Mercosur. The shaded band marks the ≥ 60° extreme regime; no real plate in any dataset falls in it, so real captures rarely exceed moderate oblique angles.

*A.1. Real-Data Recognition Validation Across CCPD and RodoSol*

This section consolidates the real-data evidence that complements the controlled synthetic grid, positioning restoration as an extreme-angle module: a standard recognizer already reads near-frontal and moderate-angle plates, so restoration is the intervention that matters once the viewing angle is oblique. Evaluated on held-out test plates (excluded from restorer training), restoration lifts recognition across CCPD and both RodoSol layouts, most strongly in the extreme regime (DS-E, > 45°), where it raises whole-plate exact-match by +15 to +38 points under the plate-specialised recognizers (Table 5) and by up to +57 points under general-purpose OCR at extreme angles (Table A1). The approximately 90–92% recoverable-grid envelope remains a controlled synthetic benchmark result; the experiments here test whether restoration improves recognition on real crops and additional plate layouts, quantify the effect by viewing-angle regime, and ablate the training regime and the restoration architecture. Section 5.6 summarizes the headline recognition result (Table 5) and the discriminative-vs-generative comparison; this appendix provides the full per-engine, per-angle-bin, and ablation detail. Because each CCPD analysis targets a different question, it draws on its own subset of the full local corpus (Appendix A.0): a low-distortion pool supplies restorer training, while each recognition table reports its own disjoint held-out test count, so the differing CCPD sample sizes are purpose-specific slices rather than conflicting totals.

Figures A1 to A3 show representative restoration examples for the three evaluated real datasets. Within each panel the left block shows real-dewarped source and restored pairs and the right block shows controlled high-angle warp/dewarp pairs. Controlled high-angle restoration recovers the plate string across all three datasets; real-dewarped recovery is partial, succeeding on the better-exposed crops, so the quantitative claims below rest on the controlled high-angle bins and the paired-CCPD evaluation.

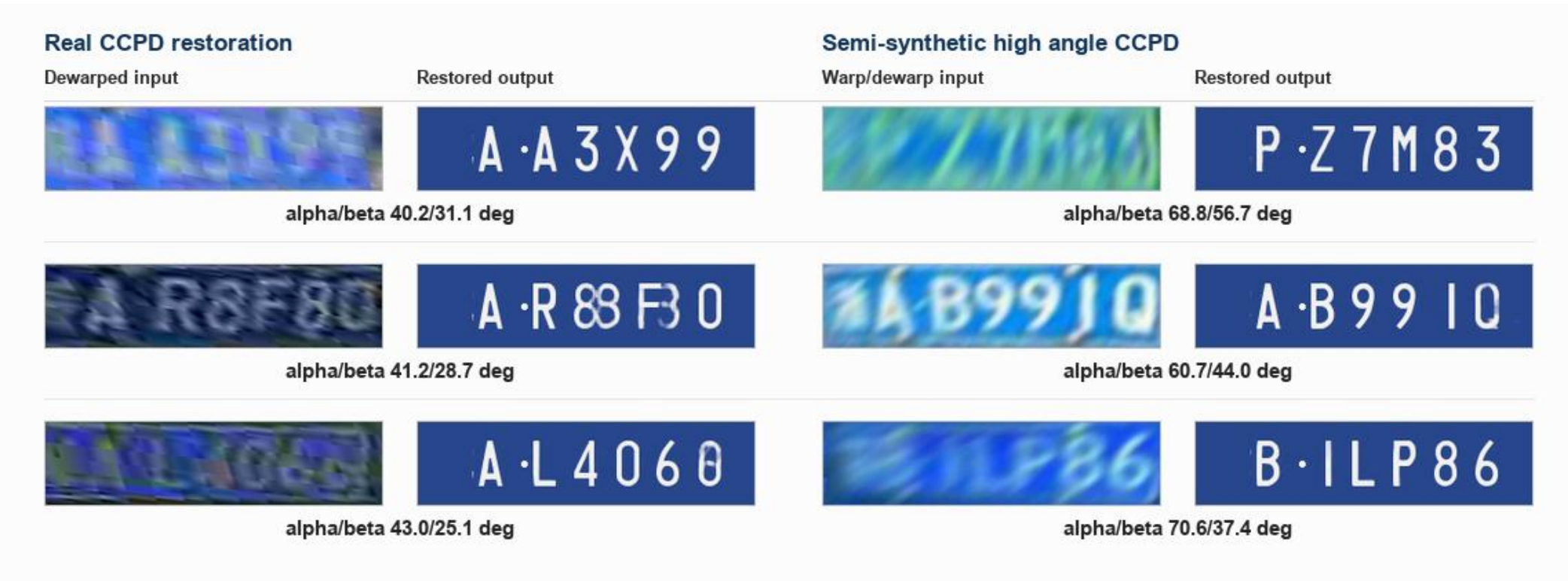

**Figure A1.** CCPD (Chinese plates) restoration examples. The left block shows three real-dewarped source/restored pairs; the right block shows three controlled high-angle warp/dewarp pairs. Real pairs are drawn from cases where the restored crop reads correctly, and controlled pairs are high-angle semi-synthetic crops with readable restored output.

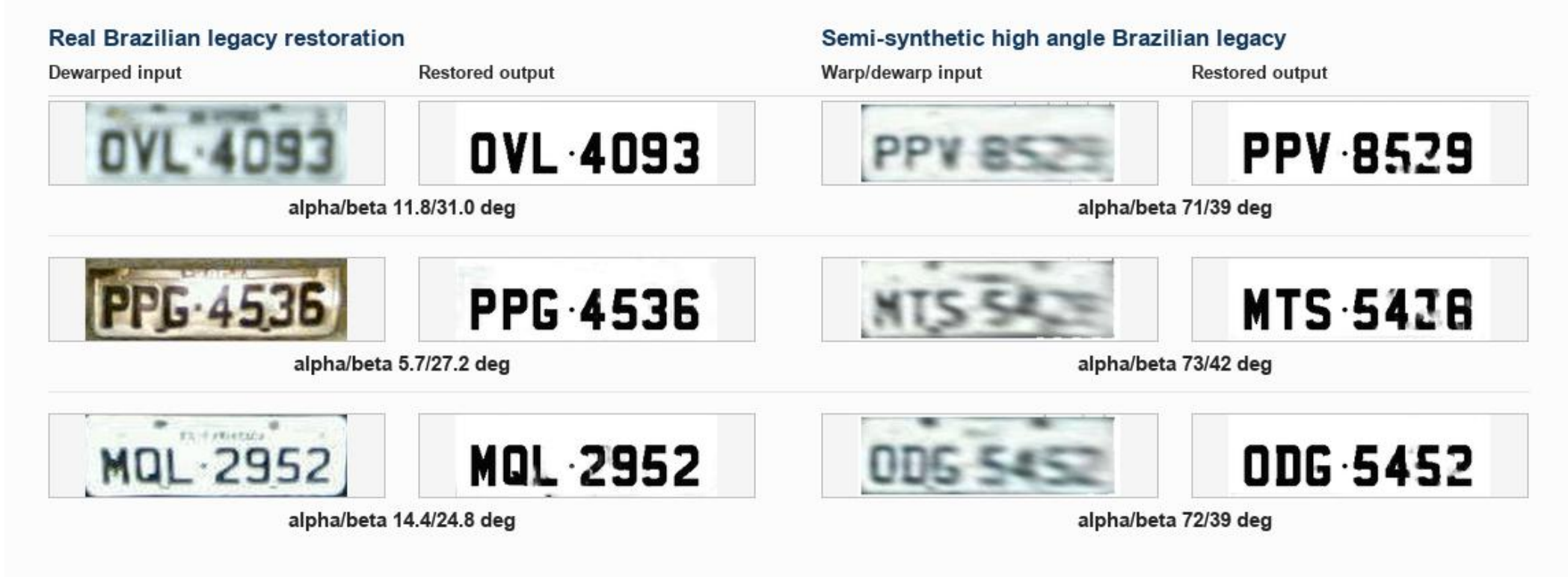

**Figure A2.** RodoSol Brazilian legacy layout restoration examples, with the same real-dewarped (left) and controlled high-angle (right) blocks as Figure A1.

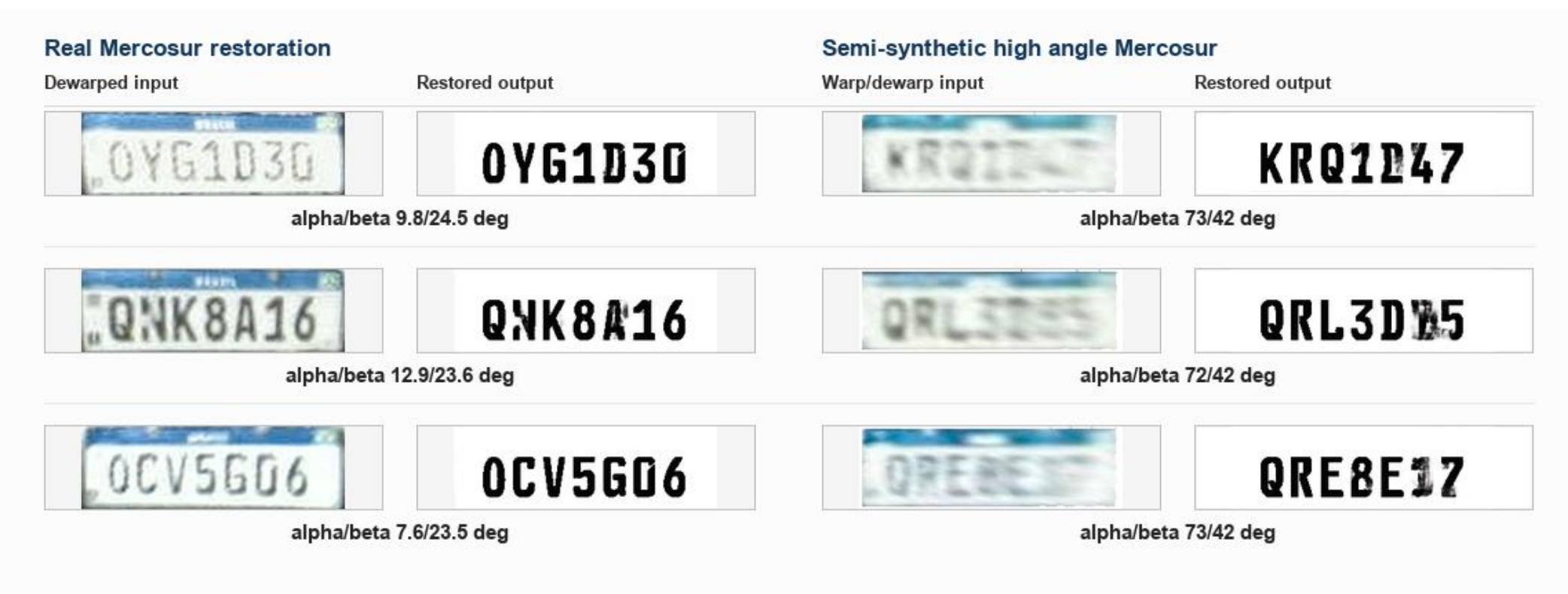

**Figure A3.** RodoSol Mercosur layout restoration examples, with the same real-dewarped (left) and controlled high-angle (right) blocks as Figure A1.

### A.1.1. CCPD paired restoration and OCR-engine sensitivity

We first evaluated whether the restoration arm improves recognition on real plates, using a labelled CCPD [25] subset with paired raw-dewarped and restored plate crops. After validity filtering, the shared sample contains 286 plates. HyperLPR3 [28] and PaddleOCR [27] are evaluated on full-plate exact match, while Tesseract [26] and EasyOCR [29] are evaluated on the alphanumeric tail because these generic OCR engines do not read the Chinese province character. Wilson 95% intervals are computed for each rate. The restored crops in this subsection are produced by a previously trained CCPD U-Net restoration checkpoint. The checkpoint is held fixed for this appendix audit: A.1.1 does not train a new network, and the target used for scoring is the labelled CCPD string rather than an image reconstruction target.

At the native, near-frontal angles of this shared real-CCPD slice (the standard regime), restoration improves every engine: HyperLPR3 full-plate exact match rises from 79.0% to 93.0%, and the general-purpose engines improve in step. The full multi-engine, per-viewing-angle-regime breakdown is reported in Table 5 (the plate-specialised HyperLPR3 and PaddleOCR recognizers, extreme regime) and Table A1 (Tesseract and EasyOCR, both regimes). The absolute rates differ by engine, which indicates that OCR choice is part of the measured system, so recoverability is reported as system-level performance under a specified OCR protocol, with multi-engine evidence as a sensitivity analysis.

#### A.1.2. Controlled high-angle CCPD evaluation

To align the CCPD [25] results with the RodoSol evaluation below, we report the controlled semi-synthetic high-angle bins at $\max(\alpha, \beta)$ equal to 50–60, 60–70, and 70–80 degrees. Low-distortion real CCPD crops are first corner-dewarped to a canonical crop. The controlled source is then generated by applying a pure high-angle warp/dewarp transform to that dewarped crop, with no added noise or blur, and the U-Net restoration is evaluated against the paired real-derived target. The metric in this panel is target-referenced PSNR, including a character-region PSNR that focuses the comparison on the license string rather than on the full plate background.

On the held-out test split, full-image PSNR improves from 10.3 to 20.3 dB in the 50–60 bin (n=9), from 10.0 to 22.0 dB in the 60–70 bin (n=12), and from 10.3 to 21.0 dB in the 70–80 bin (n=7); the panel shows 95% mean confidence intervals. These CCPD results are shown together with the RodoSol OCR results in the consolidated Figure A4, panel (a). The shared angle bins make controlled high-angle restoration directly comparable across the three real datasets without conflating PSNR evidence with OCR exact-match evidence. The CCPD DS-E recognition rates in Table A1 are computed on these same controlled high-angle bins, scored by the OCR engines rather than by PSNR.

#### A.1.3. RodoSol cross-layout real-to-clean validation

To test whether real-to-clean restoration behavior transfers beyond the single synthetic plate template, we evaluated two real Brazilian plate layouts from RodoSol-ALPR [18]: the legacy Brazilian layout and the Mercosur layout. For each layout, low-distortion real crops were corner-dewarped to a $280 \times 91$ canonical canvas. A calibrated binary black/white text-only clean target was rendered for the same plate string, with decorative bands and non-string regions removed so the supervised target is limited to the license string. Because real RodoSol captures rarely exceed moderate angles (Figure A0), the extreme regime is probed with a semi-synthetic construction: each dewarped real crop is warped to a sampled oblique angle, augmented, and dewarped back, the same dewarp-warp-augment-dewarp pipeline used to build the training data. Separate per-layout U-Nets were trained on roughly 1,750 low-distortion plates per layout, with a held-out test split reserved for evaluation, for 14 epochs with batch size 16, base width 32, learning rate $10^{-3}$, and uniform full-canvas MSE plus 0.10 L1 loss against the rendered text-only target. Recognition was scored on the held-out test plates with two real OCR engines, EasyOCR [29] and Tesseract [26], applied to the input and restored crops, with both whole-plate exact match and character accuracy ($1 - \text{CER}$) computed against the known plate string. Restoration lifts real-engine recognition across both layouts, most strongly in the extreme regime: Tesseract exact match rises from 16.6% to 73.1% for the Brazilian legacy layout in the DS-E band (max angle $> 45^\circ$, $n = 941$) and from 11.2% to 62.2% for Mercosur ($n = 941$), with character accuracy reaching 95% and 93% respectively; EasyOCR shows the same pattern (Table A1). Under each layout's plate-specialised recognizer (Fast-Plate-OCR), which reads the degraded plate on fair terms, the pooled DS-E lift is 63.2% → 86.1% (Brazilian) and 66.0% → 96.5% (Mercosur) with the best restorer (Table 5, Appendix A.1.5); Figure A4 gives the per-angle-bin breakdown of these plate-specialised rates. The standard band ($\leq 45^\circ$) and the full per-engine breakdown are in Table A1; Figure A4 shows the per-bin recognition under each dataset's plate-specialised recognizer (HyperLPR3 for CCPD, Fast-Plate-OCR for RodoSol). Whole-plate exact match at extreme angles is unforgiving, since a single mis-read glyph fails the entire plate, so the $1 - \text{CER}$ gains are the less discretised view of the same lift; both are large and consistent across layouts. Representative real and controlled high-angle restoration pairs for the two layouts appear in Figures A2 (Brazilian legacy) and A3 (Mercosur).

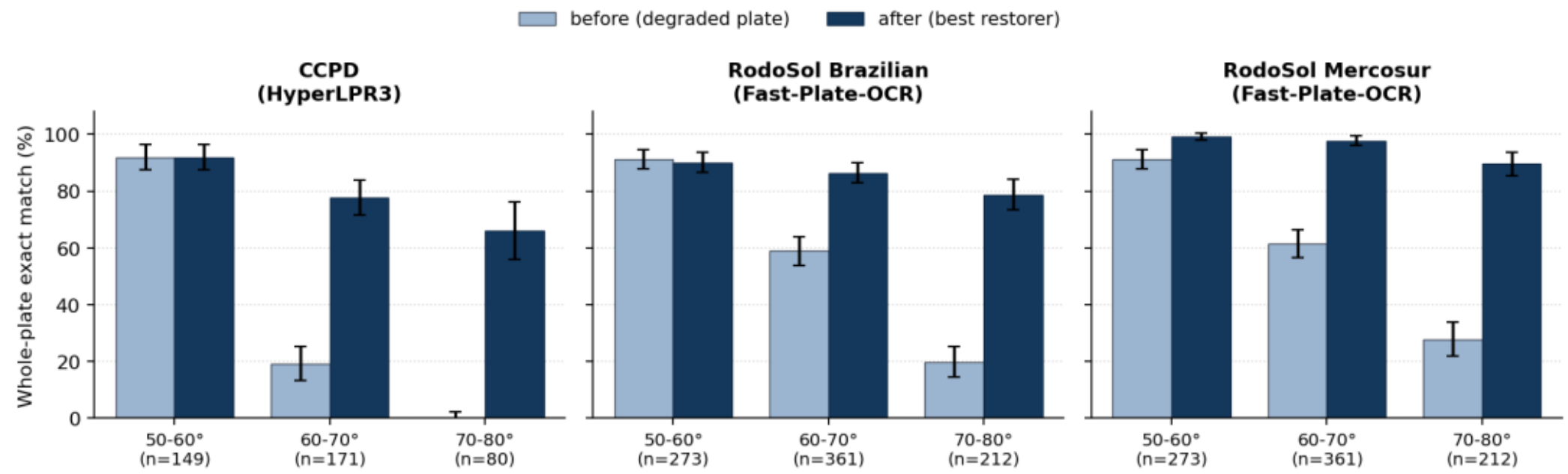

**Figure A4.** Per-bin whole-plate recognition (Wilson 95% intervals) under each dataset's plate-specialised recognizer, before (degraded crop) vs after (best restorer), on semi-synthetic crops warped to Sobol-sampled viewing angles in shared worst-axis bins (50–60, 60–70, 70–80 degrees). Panels: CCPD (HyperLPR3, U-Net), RodoSol Brazilian and Mercosur (Fast-Plate-OCR, Restormer). The gains concentrate in the 60–80 degree bins where the degraded baseline collapses; shared bins make the three plate domains comparable.

**Table A1.** General-purpose OCR (EasyOCR, Tesseract) by viewing-angle regime, before and after restoration, on held-out test plates. Each engine reads the plate's character region: for CCPD the 6-character ASCII tail (the non-Latin province character, which Latin OCR cannot read, is cropped out); for the all-Latin RodoSol plates the full plate. Each cell: exact-match (degraded → restored, with the after-rate's Wilson 95% half-interval ±) and, in parentheses, character accuracy (1 −CER); the Restorer column names the model producing the restored crop (U-Net for CCPD, Restormer for RodoSol, matching Table 5). DS-S is the standard regime (worst-axis ≤ 45°), DS-E the extreme regime (> 45°). Restoration lifts both engines most in the extreme regime; at standard angles, where the engine already reads the cropped near-frontal plate, the exact-match gain is small and for Tesseract slightly negative, while character accuracy still rises. The fair plate-specialised comparison is Table 5 (HyperLPR3 for CCPD, Fast-Plate-OCR for RodoSol).

| Dataset | Engine | Restorer | | DS-S (≤45°) | | DS-E (>45°) |
|---|---|---|---|---|---|---|
| | | | ***n*** | **degraded→restored (1−CER)** | ***n*** | **degraded→restored (1−CER)** |
| CCPD | EasyOCR | U-Net | 450 | 25.3 → **63.1** ± 4 (71→92) | 900 | 1.7 → **52.1** ± 3 (19→87) |
| CCPD | Tesseract | U-Net | 450 | 54.2 → 47.6 ± 5 (80→87) | 900 | 11.4 → **39.8** ± 3 (35→83) |
| Brazilian | EasyOCR | Restormer | 184 | 42.4 → **63.6** ± 7 (86→93) | 941 | 15.5 → **57.3** ± 3 (49→91) |
| Brazilian | Tesseract | Restormer | 184 | 45.7 → **81.5** ± 6 (80→97) | 941 | 16.6 → **73.1** ± 3 (51→95) |
| Mercosur | EasyOCR | Restormer | 184 | 12.0 → **62.0** ± 7 (74→93) | 941 | 5.1 → **54.6** ± 3 (36→91) |
| Mercosur | Tesseract | Restormer | 184 | 22.3 → **60.3** ± 7 (79→93) | 941 | 11.2 → **62.2** ± 3 (48→93) |

A detect-then-recognize recognizer must be driven through its recognition stage on pre-cropped restored plates: a plate detector rejects most clean restored crops (in our checks only about 5% are detected, versus 38% of the textured degraded inputs), because a restored crop no longer presents the bordered, textured appearance the detector expects. Called directly on the localised crops, the HyperLPR3 recognizer reads CCPD DS-E plates at 80.8% exact-match after restoration versus 42.5% on the degraded input (1 −CER 66.5% → 94.1%), and the general-purpose engines gain even more from their lower baseline (EasyOCR 0.2% → 38.0% exact-match, 1 −CER 19% → 83%). The deployment implication is to drive a specialised recognizer through its recognition stage rather than its detector when the input is an already-localised, restored plate.

A.1.4. Training-regime ablation and per-bin recoverability

A deployed pipeline can estimate the viewing angle cheaply and invoke restoration only beyond the standard regime, so we ask whether a restorer specialised to extreme angles beats a single full-range model on DS-E. We trained a DS-E-only variant (training angles > 45°) for each dataset alongside the full-range model and evaluated both on the held-out DS-E test split, restoring identical crops. The effect is small and dataset-dependent: on CCPD (drawn from ≈ 20,800 low-distortion plates) the full-range model is marginally ahead (HyperLPR3 83.7% vs 82.1% exact-match), while on the smaller RodoSol layouts the DS-E specialist gains 3–4 points (scored here with general-purpose OCR: Brazilian Tesseract 77.6% vs 73.2%; Mercosur EasyOCR 62.0% vs 58.9%). Both arms deliver the full restoration lift, so a single full-range restorer is sufficient and is the model reported throughout; specialisation helps only marginally and only on the smaller corpora. Within the extreme regime the lift grows monotonically with angle: for CCPD under HyperLPR3 restoration is neutral at 50–60° (the degraded input already reads ≈ 97% 1 −CER), rises to +44 points 1 −CER at 60–70° (48.8% → 92.8%), and +57 points at 70–80° (33.3% → 90.8%), holding character accuracy near 90% across the whole extreme band, so restoration is most valuable exactly where the unrestored plate is least readable. These per-bin rates are character accuracy (1 −CER); Figure A4 reports the same bins in whole-plate exact-match.

A.1.5. Discriminative vs generative restoration on real plates

The operational recommendation of this paper (Appendix A.2) is to prefer discriminative restorers over generative ones for identity-bearing recognition. We test it directly on real CCPD plates by training the same five architectures benchmarked on synthetic data, U-Net, U-Net Conditional, Restormer, Pix2Pix, and Diffusion-SR3, under their per-architecture recipes, then evaluating each over a dense $(\alpha, \beta)$ grid (HyperLPR3 recognition, ≥ 90% per-cell threshold) and by extreme-angle recognition across three OCR engines. Only the two discriminative convolutional and transformer restorers expand the recoverable region: U-Net most (boundary-AUC 0.445 → 0.600) and Restormer next (0.445 → 0.522), and they also lead extreme-angle exact match (HyperLPR3 84.3% and 82.8%). These five separately retrained architectures are scored on the dense $(\alpha, \beta)$ grid here; the primary U-Net reported in Table 5 reads 80.8% on the held-out DS-E test split. The Conditional U-Net and Diffusion-SR3 are flat (0.445 → 0.436 and 0.445 → 0.453), and the Pix2Pix GAN collapses coverage far below the degraded input (0.445 → 0.050, HyperLPR3 56.1%): it emits clean, confident, incorrect glyphs even on already-readable standard-angle plates, so almost no cell clears the recognition threshold. On real plates the discriminative-over-generative separation is thus visible already in coverage, a sharper result than on the synthetic benchmark, where all five architectures reach a common outer boundary (AUC 0.89 to 0.93) and separate only in interior reliability $F$. Figure A5 shows the generative failure mode on representative high-obliquity crops.

The same five-architecture comparison reproduces on the two real RodoSol layouts, scored by a plate-specialised recognizer (Fast-Plate-OCR [45]) rather than a general-purpose engine, so the degraded real plate is read on fair terms: at extreme angles the recognizer already reads the unrestored input at 63.2% (Brazilian legacy) and 66.0% (Mercosur), a far higher and more honest baseline than the general-purpose OCR of Table A1. Restoration nevertheless lifts recognition substantially under this fair engine. The discriminative restorers gain +15 to +31 exact-match points, Restormer leading on both layouts (86.1% Brazilian, 96.5% Mercosur), while the generative Diffusion-SR3 gains the least and on the Brazilian layout falls below the unrestored baseline (53.6% versus 63.2%). Scored by each dataset's plate-specialised recognizer, HyperLPR3 for CCPD and Fast-Plate-OCR for RodoSol, the ordering is therefore consistent across all three real corpora: discriminative restoration reliably recovers extreme-angle plate identity (+15 to +38 points), whereas generative restoration recovers it least and can reduce it below no restoration at all.

Together, the CCPD and RodoSol experiments provide real-data recognition checks with explicit OCR engines, angle bins, targets, training-regime and architecture ablations, and restoration protocols, while keeping the controlled synthetic benchmark claim separate from deployment transfer.

*A.2. Identity Preservation: Discriminative vs Generative Restoration*

A central operational claim of this paper is that discriminative restorers are safer than generative ones for identity-bearing recognition, because a generative model can return a clean-looking but incorrect plate. This subsection gathers the quantitative and qualitative evidence for that claim in one place. Quantitatively, the reliability score $F$ separates the two families precisely where the boundary-AUC does not: across the Standard and Extreme datasets the discriminative models stay in a narrow $F$ band (0.10 to 0.21), while Diffusion-SR3's $F$ rises from 0.57 to 1.12, even though its AUC stays within the same 0.89 to 0.93 envelope as every other model. Two restorers can thus reach the same outer recoverability boundary yet differ sharply in interior reliability: a high-AUC, high-$F$ generative model covers a wide area but contains deep interior pockets where it emits a confident wrong reading.

Figure A5 makes the mechanism visible on representative high-obliquity crops, and Figure 6 shows the same behaviour in the main qualitative panel: under low signal Diffusion-SR3 produces plausible-looking digit shapes that are not faithful to the plate identity, whereas the discriminative models degrade to a blurred, low-confidence output that a downstream confidence check can reject. The displayed outputs are drawn from the trained models evaluated in the main synthetic setting; no model is retrained for this panel. This is the identity-preservation evidence behind the recommendation to prefer discriminative restorers for automated identity decisions.

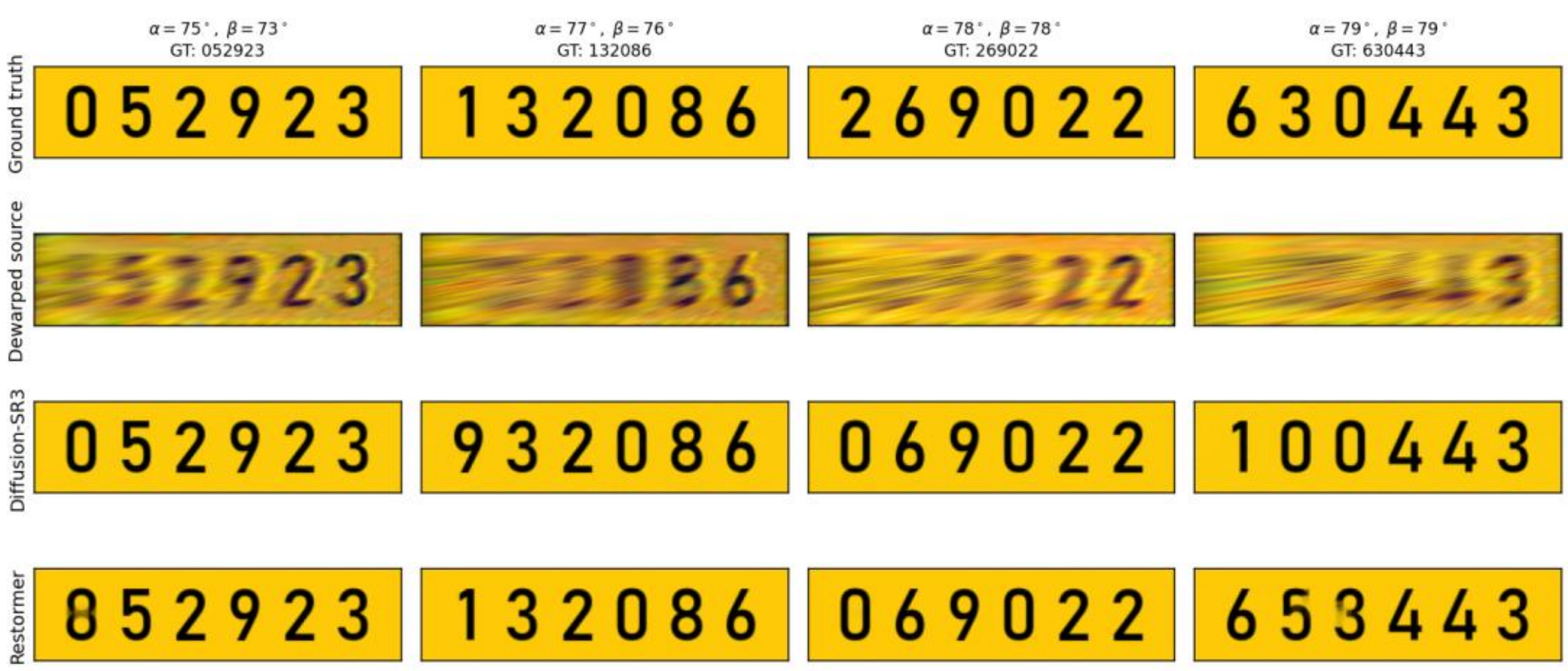

**Figure A5.** Representative qualitative examples of generative restoration artifacts. Each column shows the clean target, dewarped source input, Diffusion-SR3 output, and Restormer output. Under low signal and high obliquity, Diffusion-SR3 produces plausible-looking but incorrect digits, for example rendering the leading 1 of 132086 as a 9, whereas Restormer degrades to a rejectable blur. These models occupy the same AUC band (Table 4) yet Diffusion-SR3's reliability score $F$ is far larger, so the panel is the qualitative side of the same-AUC-different-$F$ argument and shows why interior unrecovered pockets matter for identity-recognition tasks.

### *A.3. Sobol Training-Set Size Convergence*

We trained the U-Net baseline over a Sobol sample-size sweep with $N \in \{256, 1024, 4096, 8192, 16384\}$. The trained models are evaluated with the same recoverability metrics used in the main experiment. This experiment tests whether the selected sample size is still in a steep data-scarce regime or near the plateau of the recoverability estimate. Each sample-size condition uses the same rendered clean target construction and the same Tesseract OCR scoring protocol as the main grid. The restoration model is the U-Net baseline, trained with the Section 4.3 supervised L1 plus SSIM loss. Only the Sobol training-set size is varied (Figure A6).

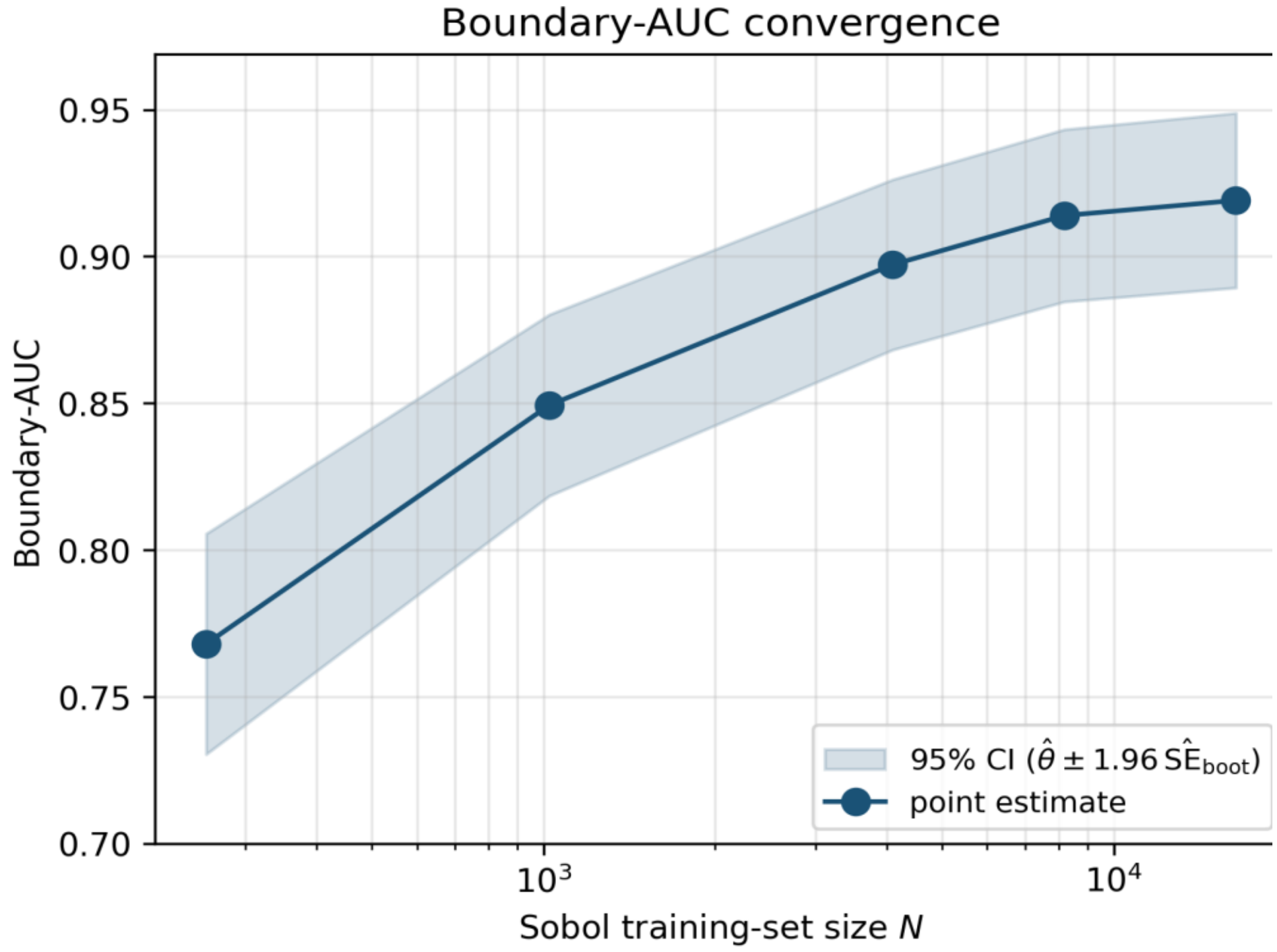

**Figure A6.** Sobol convergence of boundary-AUC for the U-Net baseline. Boundary-AUC rises from 0.768 at $N = 256$ to 0.914 at $N = 8192$ and 0.919 at $N = 16384$, indicating a plateau near the selected sampling scale.

The convergence curve reports how boundary-AUC changes with the sampling budget used in the main grid experiment. The reliability score remains more variable than AUC at small sample sizes, so fine-grained $F$ rankings are interpreted conservatively.

*A.4. Seed Stability*

The boundary-AUC 95% confidence intervals reported in Table 4 come from a stratified cluster bootstrap over grid cells (resampling rows within each $(\alpha, \beta)$ cell), which distinguishes robust coverage from fine-grained rankings: the discriminative models' AUC intervals overlap, while the reliability score $F$ is far more variable, with Diffusion-SR3 by far the widest. This is a statistical audit of existing grid outputs and does not alter training data, targets, restoration models, losses, or OCR configuration.

Beyond the grid-level bootstrap, we also audited three training seeds for the primary large-scale U-Net restoration setting (Table A2). The three runs use seeds 17, 23, and 42 and each completed 40 epochs. The best validation PSNR values are 47.47, 45.29, and 49.82 dB, with mean 47.53 dB, sample standard deviation 2.27 dB, and range 4.53 dB. The best epoch is 39 for all three runs, and final PSNR remains within 0.08 dB of the best value. The model, target construction, and OCR-independent validation metric match the primary U-Net training configuration; this audit changes only the random seed. These validation-PSNR values belong to the real-derived restoration setting and are therefore not on the same scale as the synthetic-grid test-split PSNR of Table 3, which uses a different target and degradation pipeline. The U-Net uses the same supervised L1 plus SSIM objective described in Section 4.3, and the reported metric is validation PSNR rather than a new OCR pass.

**Table A2.** Primary U-Net seed-stability audit.

| Seed | Completed epochs | Best validation PSNR (dB) | Best epoch | Final-vs-best gap |
|---|---|---|---|---|
| 17 | 40 | 47.47 | 39 | $\leq$ 0.08 dB |
| 23 | 40 | 45.29 | 39 | $\leq$ 0.08 dB |
| 42 | 40 | 49.82 | 39 | $\leq$ 0.08 dB |

This audit reports validation behavior for the primary restoration training configuration used in the supplementary real-data experiments. Architecture-level AUC/F seed matrices would provide a broader view of ranking variability across model families.